\documentclass[journal]{IEEEtran}

\usepackage{times}
\usepackage{mathrsfs}
\usepackage{epsfig}
\usepackage{graphicx}
\usepackage{amsmath}
\usepackage{amssymb}
\usepackage{graphicx}
\usepackage{cite}
\usepackage{enumerate}
\usepackage{cases}
\usepackage{multirow}
\usepackage{verbatim}
\usepackage{amssymb}
\usepackage{CJK}
\usepackage{algorithm}
\usepackage{algorithmicx}
\usepackage{algpseudocode}
\usepackage{color}
\usepackage{bm}
\usepackage{booktabs}
\usepackage[colorlinks,linkcolor=red]{hyperref}
\usepackage{amssymb}

\newcommand{\ie}{\textit{i}.\textit{e}.}
\newcommand{\eg}{\textit{e}.\textit{g}.}
\newcommand{\etc}{\textit{etc}.}
\newcommand{\wrt}{\textit{wrt}.}

\begin{document}

\title{Dense Attention Fluid Network for Salient Object Detection in Optical Remote Sensing Images}

\author
{
Qijian Zhang, Runmin Cong,~\IEEEmembership{Member,~IEEE,} Chongyi Li, Ming-Ming Cheng, Yuming Fang, Xiaochun Cao,~\IEEEmembership{Senior Member,~IEEE,} Yao Zhao,~\IEEEmembership{Senior Member,~IEEE,} and Sam Kwong,~\IEEEmembership{Fellow,~IEEE}
\thanks{Manuscript received Feb. 2020. This work was supported by the Beijing Nova Program under Grant Z201100006820016, in part by the National Key Research and Development of China under Grant 2018AAA0102100, in part by the National Natural Science Foundation of China under Grant 62002014, Grant 61532005, Grant U1936212, Grant 61971016, Grant U1803264, Grant 61922046, Grant 61772344, Grant 61672443, in part by the Hong Kong RGC General Research Funds under Grant 9042816 (CityU 11209819), in part by the Fundamental Research Funds for the Central Universities under Grant 2019RC039, in part by Elite Scientist Sponsorship Program by the Beijing Association for Science and Technology, in part by Hong Kong Scholars Program, Zhejiang Lab under Grant 2019NB0AB01, Tianjin Natural Science Foundation under Grant 18ZXZNGX00110, and in part by China Postdoctoral Science Foundation under Grant 2020T130050, Grant 2019M660438. (\emph{Qijian Zhang and Runmin Cong contributed equally to this work.) (Corresponding author: Runmin Cong.})}
\thanks{Q. Zhang is with the Department of Computer Science, City University of Hong Kong, Hong Kong SAR, China  (e-mail: qijizhang3-c@my.cityu.edu.hk).}
\thanks{R. Cong is with the Institute of Information Science, Beijing Jiaotong University, Beijing 100044, China, also with the Beijing Key Laboratory of Advanced Information Science and Network Technology, Beijing 100044, China, and also with the Department of Computer Science, City University of Hong Kong, Hong Kong SAR, China (e-mail: rmcong@bjtu.edu.cn).}
\thanks{C. Li is with the School of Computer Science and Engineering, Nanyang Technological University, Singapore  (e-mail: lichongyi25@gmail.com).}
\thanks{M.-M. Cheng is with the College of Computer Science, Nankai University, Tianjin 300071, China (e-mail: cmm@nankai.edu.cn)}
\thanks{Y. Fang is with School of Information Technology, Jiangxi University of Finance and Economics, Nanchang 330032, Jiangxi, China (e-mail: leo.fangyuming@foxmail.com).}
\thanks{X. Cao is with the State Key Laboratory of Information Security, Institute of Information Engineering, Chinese Academy of Sciences, Beijing 100093, China, and also with Cyberspace Security Research Center, Peng Cheng Laboratory, Shenzhen 518055, China, and also with School of Cyber Security, University of Chinese Academy of Sciences, Beijing 100049, China (e-mail: caoxiaochun@iie.ac.cn).}
\thanks{Y. Zhao is with the Institute of Information Science, Beijing Jiaotong University, Beijing 100044, China, also with the Beijing Key Laboratory of Advanced Information Science and Network Technology, Beijing 100044, China (e-mail: yzhao@bjtu.edu.cn).}
\thanks{S. Kwong are with the Department of Computer Science, City University of Hong Kong, Hong Kong SAR, China, and also with the City University of Hong Kong Shenzhen Research Institute, Shenzhen 51800, China (e-mail: cssamk@cityu.edu.hk).}
}

\markboth{IEEE TRANSACTIONS ON IMAGE PROCESSING}
{Shell \MakeLowercase{\textit{et al.}}: Bare Demo of IEEEtran.cls for IEEE Journals}
\maketitle

\begin{abstract}
Despite the remarkable advances in visual saliency analysis for natural scene images (NSIs), salient object detection (SOD) for optical remote sensing images (RSIs) still remains an open and challenging problem. In this paper, we propose an end-to-end Dense Attention Fluid Network (DAFNet) for SOD in optical RSIs. A Global Context-aware Attention (GCA) module is proposed to adaptively capture long-range semantic context relationships, and is further embedded in a Dense Attention Fluid (DAF) structure that enables shallow attention cues flow into deep layers to guide the generation of high-level feature attention maps. Specifically, the GCA module is composed of two key components, where the global feature aggregation module achieves mutual reinforcement of salient feature embeddings from any two spatial locations, and the cascaded pyramid attention module tackles the scale variation issue by building up a cascaded pyramid framework to progressively refine the attention map in a coarse-to-fine manner. In addition, we construct a new and challenging optical RSI dataset for SOD that contains 2,000 images with pixel-wise saliency annotations, which is currently the largest publicly available benchmark. Extensive experiments demonstrate that our proposed DAFNet significantly outperforms the existing state-of-the-art SOD competitors. \url{https://github.com/rmcong/DAFNet_TIP20}
\end{abstract}

\begin{IEEEkeywords}
Salient object detection, dense attention fluid, global context-aware attention, optical remote sensing images.
\end{IEEEkeywords}

\IEEEpeerreviewmaketitle

\section{Introduction} \label{sec1}
\IEEEPARstart{S}{ALIENT} object detection (SOD) focuses on extracting visually distinctive objects/regions from the whole field of view, which imitates the visual attention mechanism of human beings \cite{wwg_review, REVIEW, cmm_review}. Different from visual fixation prediction inspired by the gaze perception phenomena, SOD aims at completely segmenting the salient objects/regions and generating a pixel-wise saliency map \cite{Response_R4, REVIEW}. Conceptually, the processing pipeline of SOD incorporates two stages, \ie, 1) successfully determining the salient areas from background; 2) accurately segmenting the salient objects. Owing to its extendibility and efficiency, visual saliency analysis has been widely applied to a variety of down-streaming visual tasks.

Although recent decades have witnessed the remarkable success of SOD for natural scene images (NSIs), there is only a limited amount of researches focusing on SOD for optical remote sensing images (RSIs). Typically, optical RSIs cover a wide scope with complicated background and diverse noise interference. In the large-scale and challenging optical RSIs, only some local regions (\eg, man-made targets, islands, and river system) may attract humans' attention, and the SOD for optical RSIs focuses on detecting these salient targets or areas in a scene. As discussed in previous studies \cite{SGS, LVNet}, SOD for optical RSIs has extremely practical values, which has been widely used as a preprocessing technique to assist various down-streaming visual applications in the remote sensing scenes, such as image fusion \cite{Response_R9}, scene classification \cite{Response_R11}, and object detection \cite{Response_R17}. In this work, the definition of saliency inherits the concept of SOD for NSIs. Specifically, the salient objects/regions should be naturally distinct from background or related to certain object categories determined by cognitive phenomena like prior knowledge and specific tasks. It is worth mentioning that, in the context of this paper, saliency detection differs from anomaly detection \cite{anamoly_detection} that devotes to determining the abnormal objects as outliers, since abnormality does not necessarily lead to attractiveness.

\begin{figure}[!t]
\centering
\centerline{\includegraphics[width=9cm,height=5cm]{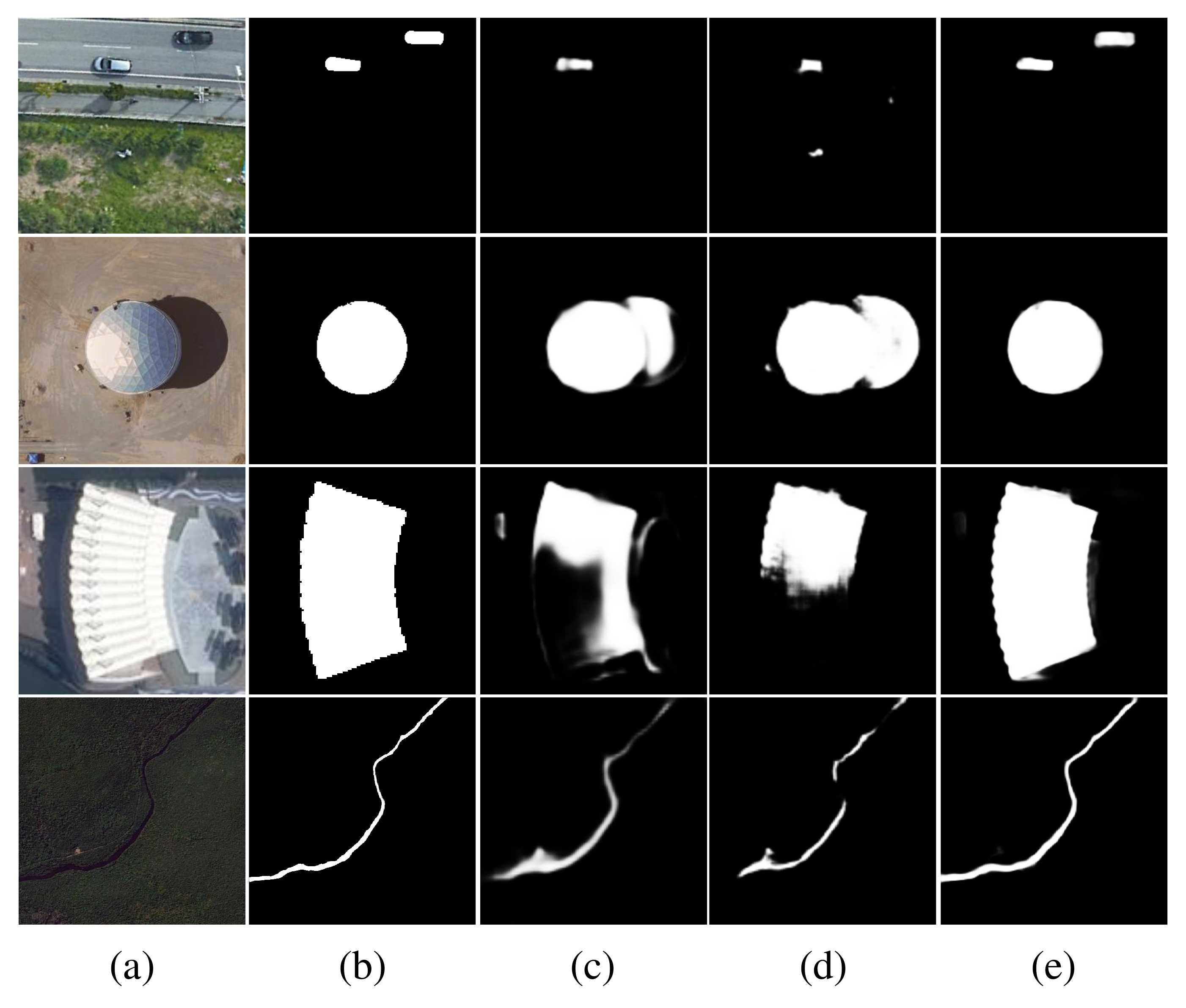}}
\caption{Visual illustration of SOD results for optical RSIs by applying different methods. (a) Optical RSIs. (b) Ground truth. (c) PFAN \cite{PFAN}. (d) LVNet \cite{LVNet}. (e) Proposed DAFNet.}
\label{fig1}
\end{figure}

Different from NSIs captured by human photographers with hand-held cameras, optical RSIs are automatically collected by various remote sensors deployed on satellites or aircrafts with minimal human intervention, which causes an obvious gap between SOD in natural and remote sensing scenes. First, RSIs are acquired through high-altitude shooting with flexible sensor-object distance, while NSIs are usually shot from a manually-adjusted and more appropriate distance. Hence, the scale of salient objects in NSIs varies within a relatively smaller range. By contrast, the salient objects appearing in RSIs show larger scale variations even for the same category. Therefore, successful NSI saliency detectors would be less reliable when dealing with the scale-variant remote sensing salient objects. Second, since RSIs are photographed from an overhead view, the included objects inevitably have various orientations. In the close-range NSIs captured from the front view, the problem of object rotation can be ignored to a large extent. Third, remote sensing scenes contain more diverse background patterns and are affected by various imaging conditions (\eg, shooting time, illumination intensity), which induces noise interference (\eg, shadows and strong exposure) and further increases the difficulty of mining saliency cues from RSIs. In summary, optical RSIs are characterized by wide coverage scope, diverse texture and visual appearance, variant object scale and orientation, and overwhelming background context redundancy. Therefore, directly applying existing SOD approaches particularly developed for NSIs to RSIs would be unreliable. This paper investigates a specialized solution to the unique and challenging problem of SOD for optical RSIs.

It is observed that SOD for optical RSIs still faces several challenges greatly hindering the detection performance. First, salient objects are often corrupted by background interference and redundancy. For example, the cluttered background in the first row of Fig. \ref{fig1} and the imaging shadows in the second row prevent existing methods (\eg, LVNet \cite{LVNet}, PFAN \cite{PFAN}) from accurately locating the salient objects, which induces larger false-negatives and false-positives, respectively. To alleviate this issue, the attention mechanism is successfully employed in SOD models to learn more discriminative features and suppress background interference  \cite{AttAdd1, AttAdd2, AttAdd4}. However, in these methods, the attentive results are independently generated from the corresponding feature levels, which is thought to be sub-optimal since they ignore the relationships among the attention maps of different levels. In fact, different levels of attention features focus on different visual contents, and these multi-level attentive cues have a positive impact on the final saliency prediction. Motivated by this, we propose a novel dense attention fluid (DAF) structure that establishes explicit connections among different levels of attention modules. In this way, low-level attentive cues can flow into deeper layers to guide the generation of high-level attention maps.

Second, salient objects in RSIs present much more complex structure and topology than the ones in NSIs, which poses new challenges for saliency detectors in capturing complete object regions. For example, the \emph{building} in the third row of Fig. \ref{fig1} cannot be detected completely by the LVNet \cite{LVNet} and PFAN \cite{PFAN} methods, and the \emph{river} with a long narrow shape in the last row tends to be broken by the existing detectors. This phenomenon is mainly caused by the feature inconsistency of distant spatial positions. In fact, previous researches have revealed that the convolution operation inevitably leads to local receptive field \cite{erf}. Common practices including multi-level intermediate feature fusion \cite{EGNet} and atrous convolution \cite{PFAN} have been exploited to overcome this limitation. However, these approaches ignore the fact that the semantic relationships among different spatial locations are also crucial for the pixel-wise SOD task. Without explicit constraints, there may still exist some significant differences between the embedded feature representations of salient pixels that are far away from each other. This intra-class feature inconsistency can cause incomplete structure of salient objects, which motivates us to propose a global context-aware attention mechanism that exploits global-context information by capturing long-range semantic dependencies among every pixel pair. The superiority of this design lies in integrating point-to-point semantic relationships to achieve feature alignment and mutual reinforcement of saliency patterns.

Third, for the optical RSI SOD task, there is only one dataset (\ie, ORSSD \cite{LVNet}) available for model training and performance evaluation, which contains $800$ images totally. This dataset is pioneering, but its data size is still relatively small. To this end, we extend it to a larger one dubbed as EORSSD including $2,000$ images and the corresponding pixel-wise saliency annotations. The EORSSD dataset is publicly available and more challenging, covering more complicated scene types, more diverse object attributes, and more comprehensive real-world circumstances.

In this paper, we devote to fully exploiting visual attention with global-context constraints for SOD in optical RSIs. The main contributions are summarized as follows:

\begin{itemize}
\item An end-to-end Dense Attention Fluid Network (DAFNet) is proposed to achieve SOD in optical RSIs, equipped with a Dense Attention Fluid (DAF) structure that is decoupled from the backbone feature extractor and the Global Context-aware Attention (GCA) mechanism.
\item The DAF structure is designed to combine multi-level attention cues, where shallow-layer attention cues flow into the attention units of deeper layers so that low-level attention cues could be propagated as guidance information to enhance the high-level attention maps.
\item The GCA mechanism is proposed to model the global-context semantic relationships through a global feature aggregation module, and further tackle the scale variation problem under a cascaded pyramid attention framework.
\item A large-scale benchmark dataset containing $2,000$ image samples and the corresponding pixel-wise annotations is constructed for the SOD task in optical RSIs. The proposed DAFNet consistently outperforms $15$ state-of-the-art competitors in the experiments.

\end{itemize}

The rest of this paper is organized as follows. In Section \ref{sec2}, we briefly review the related works of SOD in both NSIs and optical RSIs. In Sections \ref{sec3} and \ref{sec4}, we introduce the details of the proposed DAFNet and the newly constructed EORSSD benchmark dataset, respectively. In Section \ref{sec5}, the experimental comparisons and ablation analyses are discussed. Finally, the conclusion is drawn in Section \ref{sec6}.

\begin{figure*}[!t]
\centering
\centerline{\includegraphics[width=0.9\linewidth]{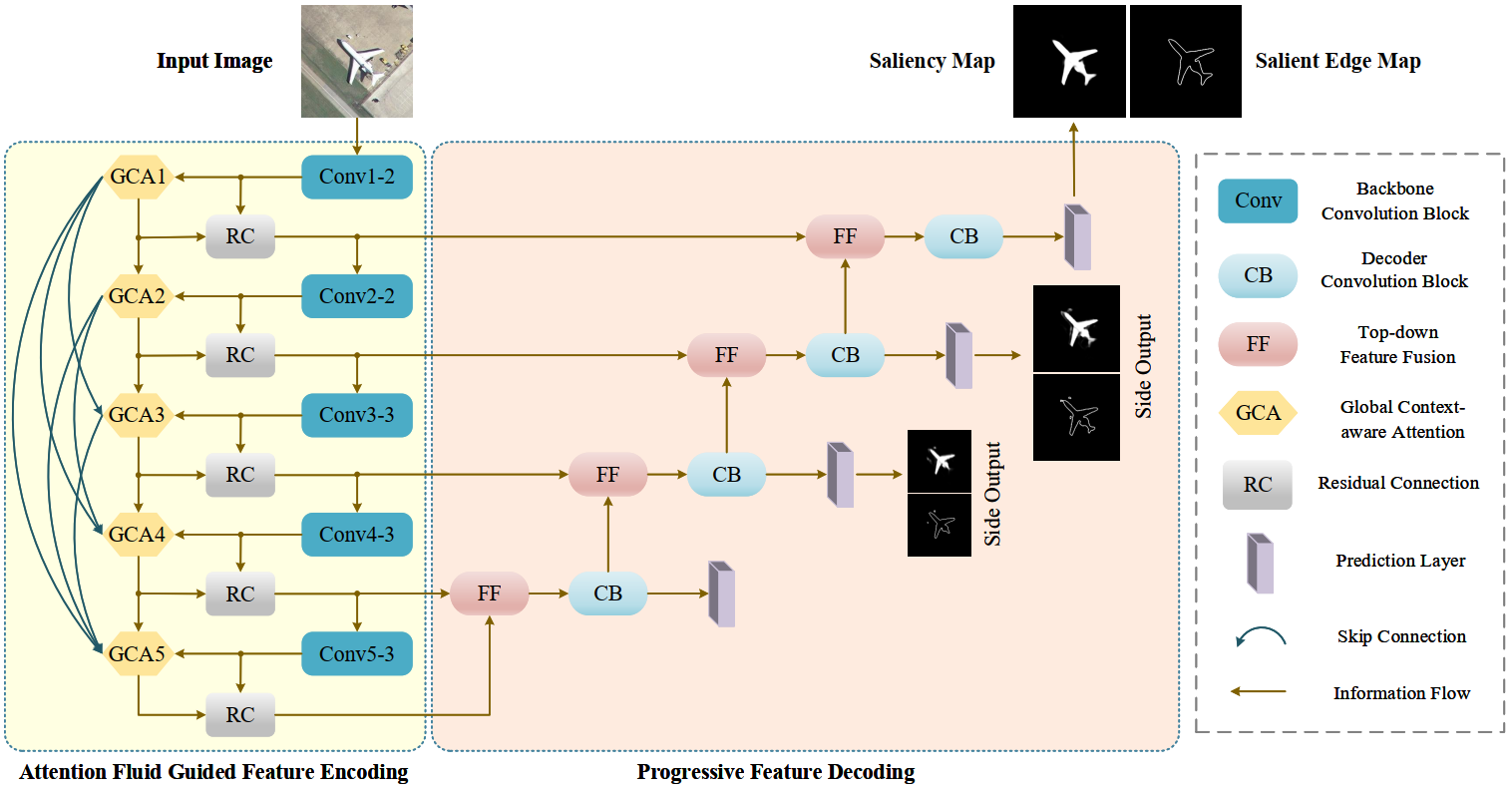}}
\caption{The overall architecture of our proposed DAFNet, including the \textbf{Attention Fluid Guided Feature Encoding} and \textbf{Progressive Feature Decoding}. The feature encoder consists of: 1) an attention fluid where low-level attention maps flow into deeper layers to guide the generation of high-level attentions, and 2) a feature fluid that generates hierarchical feature representations with stronger discriminative ability by incorporating attention cues mined from the corresponding global context-aware attention modules. The feature decoder employs a progressive top-down feature fusion strategy to produce saliency predictions at different feature scales. Note that we do not use the top-most side outputs for deep supervision in that the feature resolution is very low and hinders the depiction of detailed object structures. The \textbf{FF} unit involves up-sampling the high-level feature map and performing channel alignment via $1 \times 1$ convolutions, and then adding it to the low-level feature map. The \textbf{CB} unit is designed as a bottleneck convolution block with non-linearities to further integrate fusion feature information. The prediction layer consumes the decoded features to produce the corresponding saliency map.}
\label{fig2}
\end{figure*}
\section{Related Work} \label{sec2}
As mentioned earlier, optical RSIs have human-eye friendly color presentations similar to NSIs. However, remote sensing scenes are more challenging due to the unique imaging conditions and diverse scene patterns. In this section, we will separately review the SOD models for NSIs and optical RSIs.

\subsection{Salient Object Detection for NSIs}
The past decades have witnessed the rapid growth and flourish in model design and performance improvement of the SOD task for NSIs, especially after the rise of deep learning technology \cite{crmtc20,crmtip18,crmtc19}. Bottom-up SOD models explore various hand-crafted features or some visual priors to represent the saliency attribute, which is driven by stimulus \cite{RC,RBD,HDCT,SMD,RCRR,RRWR,DSG}. Top-down SOD models entail supervised learning with pixel-wise labels, which is driven by specific tasks, including the deep learning based methods \cite{R3Net, DSS, RADF, PFAN, EGNet, BASNet, R1_2_added_1, R1_2_added_2, R1_2_added_4, Response_R19, Response_R20, Response_R22,lcy2020tc,eccv20, PoolNet, GCPANet}. Zhao \emph{et al.} \cite{PFAN} proposed a pyramid feature attention network for saliency detection by integrating the high-level context features and the low-level spatial structural features. Hu \emph{et al.} \cite{RADF} proposed a fully convolutional neural network (FCN) based saliency network by fully exploiting the recurrently aggregated deep features captured from different layers. Liu \emph{et al.} \cite{PoolNet} introduced two simple pooling-based modules into the feature pyramid network to achieve real-time saliency detection.

\subsection{Salient Object Detection for Optical RSIs}
Compared with the research boom of SOD for NSIs, there are only three works focusing on the SOD for optical RSIs. Zhao \emph{et al.} \cite{SGS} proposed a SOD method for optical RSIs based on sparse representation using the global and background cues. Zhang \emph{et al.} \cite{SMFF} proposed a self-adaptive multiple feature fusion model based on low-rank matrix recovery for saliency detection in RSIs by integrating the color, intensity, texture, and global contrast cues. Li \emph{et al.} \cite{LVNet} proposed the first deep learning based SOD method for optical RSIs, including a two-stream pyramid module and an encoder-decoder architecture. Li \emph{et al.} \cite{lcync20} designed a parallel down-up fusion network to achieve SOD in optical RSIs.

In addition, SOD is often worked as an auxiliary component in the related optical RSI processing tasks, such as Region-of-Interest (ROI) extraction \cite{rsi1}, building extraction \cite{rsi3}, airport detection \cite{rsi4}, oil tank detection \cite{rsi5}, and ship detection \cite{rsi6}. Zhang \emph{et al.} \cite{rsi4} discovered the airport targets in optical RSIs by investigating a two-way saliency detection model that combines vision-oriented and knowledge-oriented saliency cues. Liu \emph{et al.} \cite{rsi5} proposed an unsupervised oil tank detection method that explores low-level saliency to highlight latent object areas and introduces a circular feature map to the saliency model to suppress the background. In general, since these methods are actually driven by some specific tasks, they usually show unsatisfactory performance when dealing with generic salient object detection.

\section{Proposed Method} \label{sec3}

\subsection{Overview}
In Fig. \ref{fig2}, we illustrate the overall framework of our proposed DAFNet, which is an encoder-decoder architecture. Different from conventional feature encoders where the convolutional blocks are directly stacked in sequence, we design an attention fluid to guide the backbone feature propagation, \ie, each convolutional block is equipped with a global context-aware attention (GCA) unit. For the attention information flow, we design a dense attention fluid structure, where each GCA unit produces an attention map based on raw side features generated from the corresponding convolutional block and receives the attentive results from previous GCA units (if any) to formulate a global attention map, which is further combined with raw features by means of residual connection (RC) to obtain the enhanced feature maps. Thus, in the feature fluid, we can obtain hierarchical feature representations with enhanced discriminative power after the RC operations. During feature decoding, we progressively fuse different levels of feature maps and employ additional convolutional layers to generate saliency maps and salient edge maps. In later sections, we will introduce the \emph{Attention Fluid Guided Feature Encoding}, \emph{Progressive Feature Decoding}, and \emph{Loss Function} of our proposed DAFNet.

\subsection{Attention Fluid Guided Feature Encoding}
Since the proposed pipeline of our DAFNet is independent of specific types of backbone architectures, without loss of generality, we employ the commonly used VGG16 architecture \cite{vgg} to describe our method. We remove the last three fully connected layers and truncate the first max-pooling layer in the original VGG16 network to formulate our backbone feature extractor composed of five sequentially-stacked convolutional blocks abbreviated as Conv 1-2, Conv 2-2, Conv 3-3, Conv 4-3, and Conv 5-3 in Fig. \ref{fig2}. Given an input optical RSI $\mathcal{I} \in {\mathbb{R}^{{C}\times{H}\times{W}}}$ passing through the five convolutional blocks, we obtain the raw side features at the corresponding feature levels. For simplicity, we denote the five levels of original side feature maps in a backbone feature set $f = \{ f^1, f^2, f^3, f^4, f^5 \}$, in which the numerical superscript indicates the feature level.

\subsubsection{\textbf{Global Context-aware Attention Mechanism}}
Learning more discriminative features is an essential component in salient object detection. As discussed before, in order to enable the features focus on saliency-related regions and reduce the feature redundancy, attention mechanism has been applied to SOD \cite{AttAdd4, PFAN}. However, due to the limited receptive field of convolution operations and weak constraints of feature consistency in global image context, these methods may induce incomplete saliency predictions when the foreground covers a large scope in the image. To this end, we investigate a novel global context-aware attention (GCA) mechanism that explicitly captures the long-range semantic dependencies among all spatial locations in an attentive manner. Specifically, the GCA module consists of two main functional components, \ie, global feature aggregation (GFA) and cascaded pyramid attention (CPA). The GFA module consumes the raw side features generated from the backbone convolutional block and produces aggregated features that encode global contextual information. The CPA module is used to address the scale variation of objects in optical RSIs, which takes the aggregated features from GFA as input and produces a progressively-refined attention map under a cascaded pyramid framework.

\textbf{(a) Global Feature Aggregation}. In the ideal case, for the pixels that belong to the same salient object, the learned feature is supposed to be consistent regardless of their spatial distance. However, when the salient object covers a large scope, this feature consistency becomes fragile, which leads to incomplete detection results. Thus, the GFA module aims to achieve the feature alignment and mutual reinforcement between saliency patterns by aggregating global semantic relationships among pixel pairs, which is beneficial to constrain the generation of intact and uniform saliency map.

Consider an original side feature map $f^s \in \mathbb{R}^{C_s \times H_s \times W_s}$, where $s \in \{1,2,3,4,5\}$ indexes different convolution stages. Our goal is to evaluate the mutual influence between the embedding vectors of any two positions, and then generate a global context-aware descriptor by aggregating every local feature. Mathematically, the mutual influence is formulated as a spatial correlation map $C^s \in \mathbb{R}^{P_s \times P_s}$:
\begin{equation}
C^s = {\{{(\mathcal{R}(\widetilde{f}^s))}^T \otimes \mathcal{R}(\widetilde{f}^s)\}}^T
\end{equation}
\noindent where $\widetilde{f}^s$ represents the normalized side feature, $\mathcal{R}({\cdot})$ denotes a dimension transformation operation that reshapes a matrix of $\mathbb{R}^{D_1 \times D_2 \times D_3}$ into $\mathbb{R}^{D_1 \times D_{23}}$, $D_{23}=D_2 \times D_3$, $\otimes$ is the matrix multiplication, and  $P_s={H_s \times W_s}$ counts the number of pixels. Then, we calculate a global context map $\mathcal{W}^s \in \mathbb{R}^{P_s \times P_s}$. Its $(i,j)$ entry $\omega^s_{ij}$ is defined as:
\begin{equation}
\omega^s_{ij} = \frac {e^{c^s_{ij}}} {\varphi^s_j}= \frac {e^{c^s_{ij}}} {\sum_{i=1}^{P_s} e^{c^s_{ij}}}
\end{equation}
\noindent where $c^s_{ij}$ is the entry of matrix $C^s$ at location $(i,j)$, which indicates cosine-distance based feature similarity between two embedded vectors, and $\varphi^s_j$ denotes the Gaussian weighted summation of all elements in the $j^{th}$ column of matrix $C^s$. In such context, $\omega^s_{ij}$ measures the relative impact the $i^{th}$ position has on the $j^{th}$ position, and thus global inter-pixel relationships are efficiently encoded in $\mathcal{W}^s$.

In this way, we can obtain an updated feature map $G^s$ that encodes global contextual dependencies:
\begin{equation}
G^s = \mathcal{R}^{-1}(\mathcal{R}(\widetilde{f}^s) \otimes \mathcal{W}^s)
\end{equation}
\noindent where $\mathcal{R}^{-1}$ is the inverse operator of $\mathcal{R}({\cdot})$ that reshapes a matrix of $\mathbb{R}^{D_1 \times D_{23}}$ into $\mathbb{R}^{D_1 \times D_2 \times D_3}$. Then, we integrate the updated feature into the original side feature output $f^s$ in a residual connection manner to realize feature enhancement:
\begin{equation}
\label{Eq4}
F^s = f^s + \delta \cdot (f^s \odot G^s)
\end{equation}
\noindent where $\odot$ represents element-wise multiplication, and $\delta$ is a learnable weight parameter that controls the contribution of the global contextual information. Thus, the aggregated feature map $F^s$ encodes the global contextual dependencies, which maintains the feature consistency of the whole salient regions.

In order to generate more compact feature representations, we perform feature re-calibration \cite{SE} by explicitly modeling inter-dependencies among channel responses of convolutional feature maps. Specifically, we first apply average-pooling and max-pooling operations separately over the obtained feature map $F^s$ to produce two 1D channel importance descriptors $\Gamma^s_a$ and $\Gamma^s_m$, respectively. Then, they are fed into a three-layer bottleneck fully-connected block and further combined to produce a fusion channel reweighing vector. This process can be formulated as:
\begin{equation}
\Gamma^s = \sigma(\mathcal{F}(\Gamma^s_a;\theta^s_1,\theta^s_2)+\mathcal{F}(\Gamma^s_m;\theta^s_1,\theta^s_2))
\end{equation}
\noindent where $\sigma(\cdot)$ denotes the Sigmoid activation function, and $\mathcal{F}(\Gamma^s_{a};\theta^s_1,\theta^s_2)$ means the given feature map $\Gamma^s_{a}$ passing through a three-layer bottleneck perception with the weights $\theta^s_1$ and $\theta^s_2$. Since $\Gamma^s$ encodes cross-channel correlations and highlights the most important feature channels, we multiply it with $F^s$ by broadcasting along the spatial dimension to generate a refined feature map $F^s_{g}$ with more compact channel information.

\begin{figure}[!t]
\centering
\centerline{\includegraphics[width=8.7cm,height=6cm]{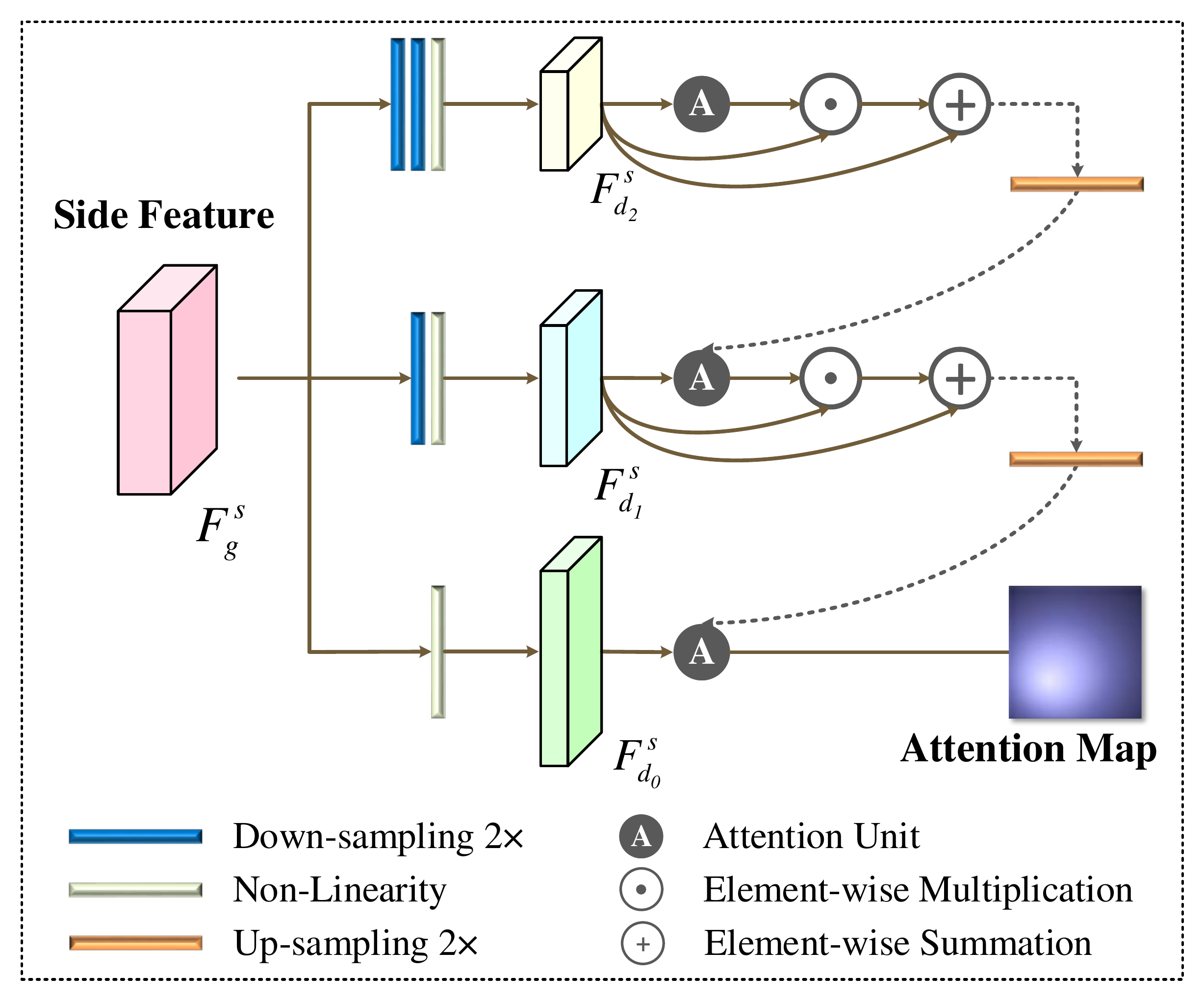}}
\caption{Illustration of the proposed CPA module, in which the pyramid attention is cascaded to progressively refine both features and attentive cues from coarse to fine.}
\label{fig3}
\end{figure}

\textbf{(b) Cascaded Pyramid Attention}. In the optical RSIs, the object scale varies greatly, which challenges the robustness and generalization of SOD models. As explored in previous works \cite{AttAdd4, PFAN}, multi-scale pyramid attention architecture enhances the discrimination of features, but also resists the size change of the objects to a certain extent. However, it would be sub-optimal to integrate the attention maps generated independently at different feature resolutions simply through up-sampling and point-wise summation, mainly because such a design weakens the information interaction between different scales and limits the expression ability of the multi-scale features. Hence, we design a cascaded pyramid attention module to progressively refine both features and attentive cues from coarse to fine. An illustration of the CPA module is provided in Fig. \ref{fig3}. To facilitate description, we begin with inferring a single-scale attention map from a given feature map by employing an efficient spatial attention approach \cite{cbam}. Specifically, we apply average-pooling and max-pooling to $F^s_{g}$ along the channel axis, and concatenate the outputs to generate a spatial location attentive descriptor $\Omega^s \in \mathbb{R}^{2 \times H_s \times W_s}$, which is further convolved to produce a 2D spatial attention map $A^s \in \mathbb{R}^{H_s \times W_s}$:
\begin{equation}
\begin{aligned}
A^s &= Att(F^s_{g})=\sigma(conv(\Omega^s);\hat{\theta}))\\
&=\sigma(conv(concat(avepool(F^s_{g}),maxpool(F^s_{g}));\hat{\theta}))
\label{eq}
\end{aligned}
\end{equation}
\noindent where $conv(\cdot ;\hat{\theta})$ denotes a customized convolutional layer with the parameter $\hat{\theta}$, $avepool$ and $maxpool$ are the average-pooling and max-pooling, respectively, and $concat$ represents feature channel concatenation.

In order to obtain multi-scale pyramid features, we first progressively down-sample the feature map $F^s_{g}$ into different resolutions using stacked $2\times$ max-pooling operations, and squeeze the feature channels with by $1\times1$ convolutions. Thus, we construct a feature pyramid $F^s_{d_k} \in \mathbb{R}^{\frac{C_{s}}{3} \times {\frac{H_s}{2^{k}}} \times {\frac{W_s}{2^{k}}}}$, where $k \in \{ 0,1,2 \}$ indexes the pyramid scale. Then, starting from the lowest resolution (\ie, $4\times$ down-sampling), we infer an attention map $A^s_{d_2}$ by Eq. \ref{eq}. The attentive cue mined at this scale is applied over the corresponding feature, which is further propagated to the next scale for attention generation. The attention map at the middle feature resolution (\ie, $2\times$ down-sampling) is generated by:
\begin{equation}
A^s_{d_1} = Att(concat(F^s_{d_1},(F^s_{d_2} \circledcirc A^s_{d_2} + F^s_{d_2})\uparrow))
\end{equation}
\noindent where $\circledcirc$ denotes element-wise multiplication with channel-wise broadcasting, and $\uparrow$ means the $2\times$ spatial up-sampling operation. By analogy, the CPA module will produce a full-resolution attention map $\hat{A^s}$ at the original feature scale, which can be formulated as:
\begin{equation}
\hat{A}^s = Att(concat(F^s_{d_0},(F^s_{d_1} \circledcirc A^s_{d_1} + F^s_{d_1})\uparrow))
\end{equation}
Through the cascaded architecture, the coarse attention cues are mined based on lower-resolution features, then combined with higher-resolution features to generate fine attention results with more accurate details.

\begin{figure*}[!t]
\centering
\centerline{\includegraphics[width=1\linewidth]{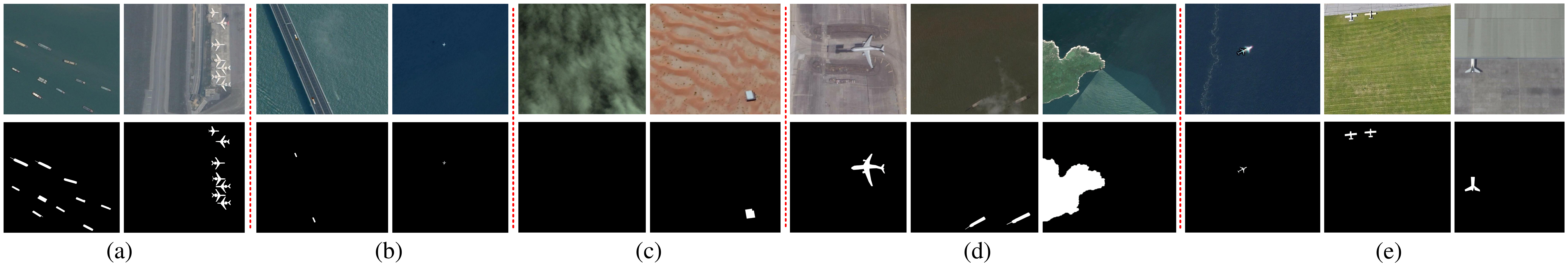}}
\caption{Visualization of the more challenging EORSSD dataset. The first row shows the optical RSI, and the second row exhibits the corresponding ground truth. (a) Challenge in the number of salient objects. (b) Challenge in small salient objects. (c) Challenge in new scenarios. (d) Challenge in interferences from imaging. (e) Challenge in specific circumstances.}
\label{fig4}
\end{figure*}
\subsubsection{\textbf{Dense Attention Fluid Structure}}

Hierarchical feature representations within convolutional networks naturally correspond to different levels of information abstraction. Shallow features focus on edges and unique textures, while deep features capture high-level visual semantics. In fact, the attentive cues derived from different convolutional stages also imply different feature selection strategies. Motivated by the improvement brought by side-path connections of convolutional features, we build a novel DAF structure where shallow-layer attention information flows into the attention units of deep layers. In this way, low-level attentive cues could be propagated as useful guidance information to enhance and refine high-level attention generation.

As a review, each GCA module consumes a raw side feature map $f^s$ to produce an attention map $\hat{A}^s$. First, we build sequential connections among the attention maps generated from the hierarchical feature representations. Moreover, considering the hierarchical attention interaction among different levels, we add feed-forward skip connections to form the attention fluid. In order to update $\hat{A}^s$, the down-sampled attention maps $\{ \hat{A}^1,\hat{A}^2,...,\hat{A}^{s-1} \}$ from shallow layers are firstly concatenated with $\hat{A}^s$ along the channel dimension. After that, a convolution layer followed by a Sigmoid function is employed to generate the final attention map. Formally, the above updating process is denoted as:
\begin{equation}
\hat{A}^s \gets \sigma (conv(concat((\hat{A}^1)\downarrow,...,(\hat{A}^{s-1})\downarrow,\hat{A}^s)))
\end{equation}

\noindent where operator $\downarrow$ means down-sampling the given attention map with higher resolution to the same size of $\hat{A}^s$.

With the updated attention map, the final feature map at the $s^{th}$ convolution stage $F^s_c$ can be generated via the residual connection:
\begin{equation}
F^s_c = concat(F^s_{d_0},(F^s_{out_1})\uparrow) \circledcirc (\hat{A}^s + O^s)
\end{equation}

\noindent where $F^s_{out_1}=F^s_{d_1} \circledcirc A^s_{d_1} + F^s_{d_1}$ is the attention weighted feature output at scale $1$, and $O^s \in \mathbb{R}^{H_s \times W_s}$ is a matrix with all entries equal to $1$. Thus, we could sequentially obtain the corresponding enhanced side features $\{ F^1_c,F^2_c,F^3_c,F^4_c,F^5_c \}$ from the five convolution stages.

\subsection{Progressive Feature Decoding}
In the feature decoder, we fuse deep features with shallow features progressively to produce multiple side outputs at different feature resolutions. Each decoding stage consists of three procedures, as shown in Fig. \ref{fig2}. First, we employ top-down feature fusion (FF) to align the spatial resolution and number of channels between adjacent side feature maps via up-sampling and $1 \times 1$ convolution, and then perform point-wise summation. Second, a bottleneck convolutional block (CB) is deployed to further integrate semantic information from fusion features. In our design, each CB unit contains two convolutional layers where the number of feature channels is first halved and then restored. Third, we deploy a mask prediction layer and an edge prediction layer for the decoded features, and use a Sigmoid layer to map the range of saliency scores into $[0,1]$. Depending on feature channel quantities of different decoding scales, the prediction layer transforms 3D features into 2D saliency map and salient edge map. The final output of our DAFNet is derived from the predicted saliency map at the top decoding level.

\subsection{Loss Function}
To accelerate network convergence and yield more robust saliency feature representations, we formulate a hierarchical optimization objective by applying deep supervisions to the side outputs at different convolution stages. Inspired by \cite{AttAdd4}, we further introduce edge supervisions to capture fine-grained saliency patterns and enhance the depiction of object contours.

\textbf{Saliency Supervision.} Given a predicted saliency map $\mathcal{S}_m$ at a certain stage and the corresponding ground truth $\mathcal{L}_m$, we employ a class-balanced binary cross-entropy loss function:
\begin{equation}\label{Eq11}
{\ell}_m = -[{\alpha}_m \cdot \mathcal{L}_m log(\mathcal{S}_m) + {\beta}_m \cdot (1-\mathcal{L}_m) log(1-\mathcal{S}_m)]
\end{equation}
\noindent where ${\alpha}_m=(\mathcal{B}-\mathcal{B}_m)/\mathcal{B}$ and ${\beta}_m=\mathcal{B}_m/\mathcal{B}$ balance the contribution of salient and background pixels. Here, $\mathcal{B}_m$ counts the positive pixels in $\mathcal{L}_m$ and $\mathcal{B}$ counts the total pixels.

\textbf{Salient Edge Supervision.} Given a predicted salient edge map $\mathcal{S}_e$, we employ the \textit{Canny} operator to extract edge pixels from $\mathcal{L}_m$, which are further strengthened into two-pixel-thick edges $\mathcal{L}_e$ through $2 \times 2$ maximum filtering. Similar to Eq. (\ref{Eq11}), we formulate the salient edge loss function as follows:
\begin{equation}
{\ell}_e = -[{\alpha}_e \cdot \mathcal{L}_e log(\mathcal{S}_e) + {\beta}_e \cdot (1-\mathcal{L}_e) log(1-\mathcal{S}_e)]
\end{equation}
\noindent where we follow the same procedure as mentioned above to compute the balancing factors ${\alpha}_e$ and ${\beta}_e$.

\textbf{Overall Loss.} Three levels of side supervisions are involved to fully exploit multi-scale information. In addition to the full-resolution results at stage $1$, we continue to make predictions at deeper stages of $2$ and $3$. We ignore the predictions with the lowest resolution since the results are very rough. Formally, the overall loss function is formulated as follows:
\begin{equation}
\label{Eq16}
{\ell} = \sum_{s=1}^3 {(w_m \cdot {\ell}_m^s + w_e \cdot {\ell}_e^s)}
\end{equation}
\noindent where $w_m$ and $w_e$ control the contribution of saliency supervisions and salient edge supervisions. ${\ell}^s_m$ and ${\ell}^s_e$ represent the loss functions computed at stage $s$.

\begin{figure*}[!t]
\centering
\centerline{\includegraphics[width=1\linewidth]{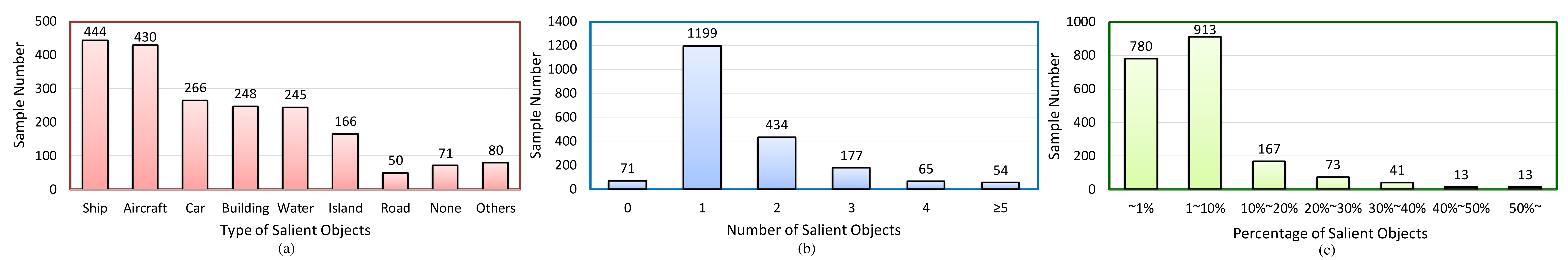}}
\caption{Statistical analysis of EORSSD dataset. (a) Type analysis of salient object. (b) Number analysis of salient object. (c) Size analysis of salient object.}
\label{data}
\end{figure*}
\section{Extended Optical Remote Sensing Saliency Detection (EORSSD) Dataset} \label{sec4}

\subsection{Overview}
In literature \cite{LVNet}, an optical remote sensing saliency detection (ORSSD) dataset\footnote{ \url{https://li-chongyi.github.io/proj_optical_saliency.html}} with pixel-wise ground truth is built, including $600$ training images and $200$ testing images. This is the first publicly available dataset for the RSI SOD task, which bridges the gap between theory and practice in SOD for optical RSIs, but the amount of data is still slightly insufficient to train a deep learning based model. To enlarge the size and enrich the variety of the dataset, we extend our ORSSD dataset to a larger one named Extended ORSSD (EORSSD) dataset with $2,000$ images and the corresponding pixel-wise ground truth, which includes many semantically meaningful but challenging images. Based on the ORSSD dataset, we collect additional $1,200$ optical remote sensing images from the free Google Earth software, covering more complicated scene types, more challenging object attributes, and more comprehensive real-world circumstances. During the dataset construction, we follow the previous labeling protocol as proposed in \cite{Response_R5}. Specifically, for labeling the ground truth saliency masks, in our project, we invited 9 researchers with relevant professional backgrounds as our annotators and asked them to independently indicate which parts of the image they thought were visually-attractive. Based on the records of all annotators' decisions, we picked out as salient the commonly-agreed regions and objects that are voted by at least half of the annotators. After that, we carefully generate pixel-wise saliency masks for these selected regions using Photoshop. For clarity, the EORSSD dataset is divided into two parts, \ie, $1,400$ images for training and $600$ images for testing. Some illustrations of the EORSSD dataset are shown in Fig. \ref{fig4}.

\subsection{Dataset Statistics and Challenges}
In this section, we illustrate the challenges of EORSSD dataset by providing some visual samples in Fig. \ref{fig4} and statistical analysis in Fig. \ref{data}.
\begin{itemize}
\item \textbf{Challenge in the number of salient objects.} In the EORSSD dataset, more often than not, there are multiple salient objects in one image. As shown in Fig. \ref{fig4}(a), there are eleven ships in the first image, and seven airplanes in the second image. In Fig. \ref{data}(b), we count the number of salient objects in the dataset, and scenarios with more than two targets account for $36.5\%$ of the total.

\item \textbf{Challenge in small salient objects.} Target size in optical RSIs is often very diverse due to the satellite- and airborne-derived imaging. The resulting small object detection problem is also a very common but challenging problem. In the the EORSSD dataset, we collected a number of small samples, such as the vehicle in the first image of Fig. \ref{fig4}(b), and the aircraft flying through the sea in the second image of Fig. \ref{fig4}(b). In Fig. \ref{data}(c), we count the size of the salient objects in the dataset (\ie, the proportion of the salient object in the image), and the scenes with a proportion of less than $10\%$ account for $84.65\%$ of the total. Furthermore, in $39\%$ of the scenes, the salient objects occupy less than $1\%$ of the image, which illustrates the challenge of the EORSSD dataset.

\item \textbf{Challenge in more abundant scenarios.} In the EORSSD dataset, we further enriched the type of scene and also increased the difficulty of the scene, such as the cloud map in the first image of Fig. \ref{fig4}(c), and buildings in the desert in the second image of Fig. \ref{fig4}(c). In Fig. \ref{data}(c), we show the main type of salient objects, including ship, aircraft, car, building, water, island, road, \etc. Among them, ship and aircraft accounted for the higher proportion, \ie, $22.2\%$ and $21.5\%$, respectively.

\item \textbf{Challenge in various imaging interferences.} In the EORSSD dataset, we added some interference cases due to the imaging reasons, which is widespread in practical applications. For example, as shown in Fig. \ref{fig4}(d), the aircraft in the first image is distorted, the ship in the second image is occluded by clouds, and the illumination distortion occurs in the third image.

\item \textbf{Challenge in specific circumstances.} In order to increase the diversity of data samples, we collected some specific circumstances in the EORSSD dataset. For example, in Fig. \ref{fig4}(e), the aircraft on the sea with over-exposure in the first image, the aircraft on the lawn in the second image, and the aircraft in the cabin that only the tail is visible in the third image.

\end{itemize}

In summary, the new EORSSD dataset is much larger, more comprehensive, more diverse, more challenging, more capable of handling practical problems, and more conducive to the training of deep learning based SOD models. Moreover, this dataset is available at our project website: \url{https://github.com/rmcong/EORSSD-dataset}\footnote{According to the relevant data acquisition and redistribution polices of Google Earth, our dataset can only be used for academic purposes.}.\par

\section{Experiments} \label{sec5}

\subsection{Evaluation Metrics}

For quantitative evaluations, we employ the Precision-Recall (P-R) curve, F-measure, MAE score, and S-measure to evaluate the SOD performance. First, we threshold the generated saliency maps into some binary saliency maps by using a series of fixed integers from $0$ to $255$. Then, the precision and recall scores can be further calculated by comparing the binary mask with the ground truth. Taking the the precision score as the vertical axis, and the recall score as the horizontal axis, we can draw the P-R curve under different combination of precision and recall scores \cite{crmtip19,crmspl}. The closer the P-R curve is to the coordinates $(1,1)$, the better the performance achieves. F-measure is defined as a weighted harmonic mean of precision and recall, which is a comprehensive measurement, and the larger the better \cite{Fmeasure2,nips20}, \ie, $F_{\beta}=\frac{(1+\beta^{2})Precision\times Recall}{\beta^{2}\times Precision+ Recall}$, where $\beta^{2}$ is set to $0.3$ for emphasizing the precision as suggested in \cite{Fmeasure2,crmicme}. The larger $F_{\beta}$ value indicates the better comprehensive performance. Comparing the continuous saliency map $S$ with ground truth $G$, the difference is defined as MAE score \cite{MAE,crmtmm19}, \ie, $MAE =\frac{1}{w\times h} \sum_{i=1}^{w} \sum_{j=1}^{h} |S(i,j)-G(i,j)|$, where $w$ and $h$ denote the width and height of the testing image, respectively. The smaller the MAE score is, the better performance achieves. S-measure describes the structural similarity between the saliency map and ground truth, and the larger value means better performance \cite{S-measure}, \ie, $S_m = \alpha \times S_o+(1-\alpha) \times S_r$, where $\alpha$ is set to $0.5$ for assigning equal contribution to both region $S_r$ and object $S_o$ similarity as suggested in \cite{S-measure}.
\begin{figure*}[!t]
\centering
\centerline{\includegraphics[width=18.5cm,height=11cm]{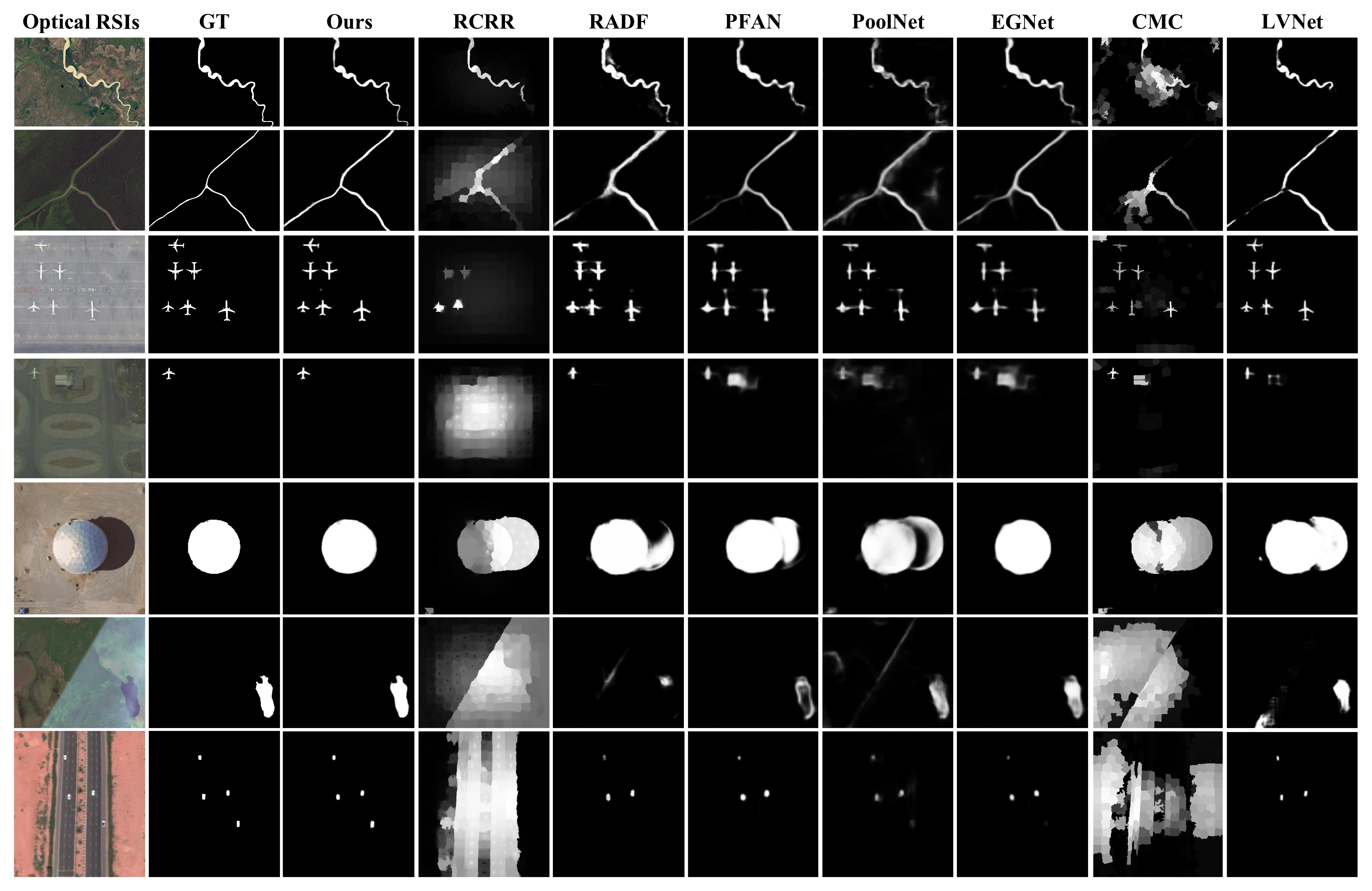}}
\caption{Visual comparisons of different methods. The proposed method under the VGG16 backbone and deep learning based methods are trained/re-trained on the EORSSD dataset.}
\label{fig5}
\end{figure*}

\subsection{Training Strategies and Implementation Details}

\textbf{Network Training.} We randomly selected $1400$ images from the constructed EORSSD dataset for training and the rest $600$ images as the testing dataset. Data augmentation techniques involving combinations of flipping and rotation are employed to improve the diversity of training samples, which produces seven variants for every single sample. Consequently, the augmented training set provides $11,200$ pairs of images in total. During training stages, the samples are uniformly resized to $128 \times 128$ due to the limited availability of our computational resources.

\textbf{Implementation Details.} We implemented the proposed DAFNet with Pytorch on a workstation equipped with an NVIDIA GeForce RTX 2080Ti GPU. The network is trained using ADAM \cite{adam} optimization strategy for $60$ epochs, and the batch size is set to $8$. In the training phase, the learning rate is fixed to $1e^{-4}$ for the first $20$ epochs, and then evenly decline to $1e^{-6}$. Filter weights were initialized by Xavier policy \cite{Xavier} and the bias parameters were initialized as constants. In the last $10$ epochs, we activate the online hard example mining \cite{ohem} mechanism where half of a batch data with lower loss values are dropped. The learnable weight $\delta$ in the GFA module is initialized as $0.1$, and the weight parameters $w_m$ and $w_e$ in loss function are set to $0.7$ and $0.3$, respectively. The average running time for processing an image with the size of $128 \times 128$ is about $0.038$ second.

\begin{figure}[!t]
\centering
\centerline{\includegraphics[width=1\linewidth]{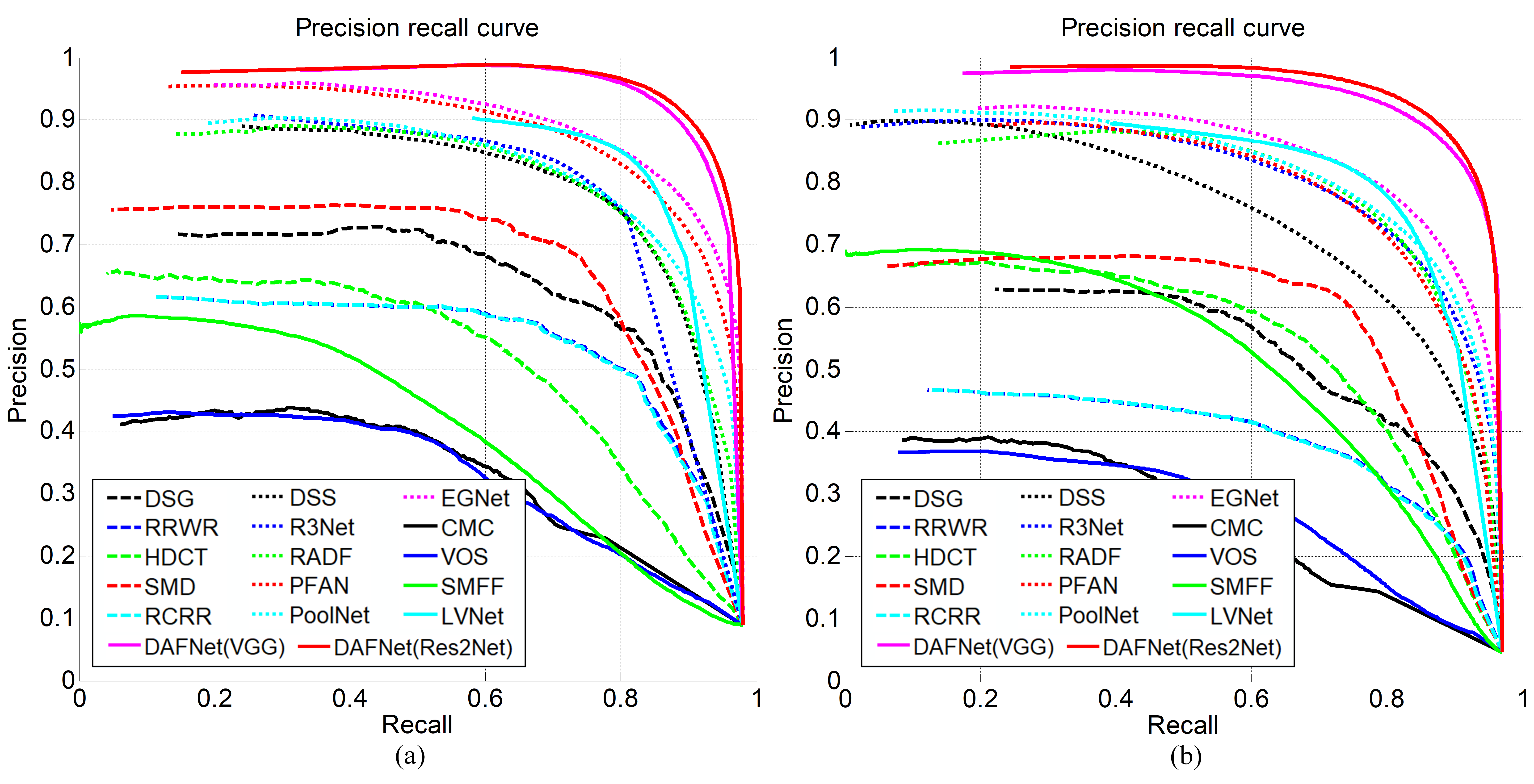}}
\caption{P-R curves of different methods on the testing subset of the ORSSD and EORSSD datasets. (a) P-R curves on the testing subset of the ORSSD dataset. (b) PR curves on the testing subset of the EORSSD dataset.}
\label{fig6}
\end{figure}
\subsection{Comparison with State-of-the-art Methods}

To verify the effectiveness of the proposed method, we compare it with fifteen state-of-the-art SOD methods on the testing subset of ORSSD dataset and EORSSD dataset, including five unsupervised methods for NSIs (\ie, RCRR \cite{RCRR}, HDCT \cite{HDCT}, SMD \cite{SMD}, RRWR \cite{RRWR}, and DSG \cite{DSG}), six deep learning-based methods for NSIs with retraining (\ie, R3Net \cite{R3Net}, DSS \cite{DSS}, RADF \cite{RADF}, PFAN \cite{PFAN}, PoolNet \cite{PoolNet}, and EGNet \cite{EGNet}), and four methods for optical RSIs (\ie, SMFF \cite{SMFF}, VOS \cite{rsi4}, CMC \cite{rsi5}, and LVNet \cite{LVNet}). All the results are generated by the source codes or provided by the authors directly, and the deep learning based methods (\ie, R3Net \cite{R3Net}, DSS \cite{DSS}, RADF \cite{RADF}, PFAN \cite{PFAN}, PoolNet \cite{PoolNet}, EGNet \cite{EGNet}, and LVNet \cite{LVNet}) are retrained by using the same training data as the proposed method under the default parameter settings in the corresponding models.

\subsubsection{\textbf{Qualitative Comparison}}
The visual comparisons of different methods are shown in Fig. \ref{fig5}, where the proposed method under the VGG16 backbone and deep learning based methods are trained/retrained on the EORSSD dataset. In Fig. \ref{fig5}, we show five challenging scenes including the river, aircraft, building and car. Compared with other methods, the proposed method exhibits superiority and competitiveness in the following aspects:

\textbf{(a) Advantages in location accuracy.} Our method can accurately locate the salient objects and has a superior ability to suppress the background interference. In the second image, the unsupervised SOD methods (\ie, RCRR \cite{RCRR} and CMC \cite{rsi5}) can hardly detect the valid salient regions, and deep learning based methods including the LVNet \cite{LVNet} can only roughly detect some parts of the salient objects. By contrast, the river is accurately and completely detected by our method.

\textbf{(b) Advantages in content integrity.} In the first image, the river has many curved structural details and relatively thin tributaries, thus some methods fail to completely detect them (\eg, PFAN \cite{PFAN}, PoolNet \cite{PoolNet}, EGNet \cite{EGNet}, and LVNet \cite{LVNet}). In the third image, all these six aircrafts should be detected as salient objects. However, the unsupervised methods (\ie, RCRR \cite{RCRR} and CMC \cite{rsi5}) cannot effectively detect all of them. Even for the deep learning based methods, although they can roughly locate the positions of all aircraft, they cannot guarantee the structural integrity of the detected aircraft. For example, both the wing and tail are lost in the results produced by the PFAN \cite{PFAN} and EGNet \cite{EGNet} methods. By contrast, our proposed method achieves better performance in structural integrity and quantitative integrity.

\textbf{(c) Advantages in challenging scenarios.} For the small and multiple objects scenarios, our algorithm still achieves competitive performance. For example, in the last image, four cars are accurately detected by our method, while other methods almost missed the car in the lower right. Similarly, in the fourth image, other methods cannot clearly and completely detect this small aircraft, or even the background regions are wrongly preserved (\eg, PFAN \cite{PFAN}, EGNet \cite{EGNet}, and LVNet \cite{LVNet}), while our method addresses these aspects very well. In addition, the proposed method is more robust to the noises from the imaging and shadow. For example, in the fifth image, only the proposed method and EGNet \cite{EGNet} method successfully suppress the shadow area, while other methods detect it as the salient region without exception. Moreover, compared with the EGNet method, our method obtains more accurate structure information and better boundary retention ability. In the sixth image, even in the case of drastic changes in lighting, our method can still accurately highlight the salient objects and effectively suppress the background.

\begin{table}[!t]
\renewcommand\arraystretch{1.2}
\caption{Quantitative Comparisons with Different Methods on the Testing Subset of the ORSSD and EORSSD Datasets. Top Three Results Are Marked in \textcolor[rgb]{1.00,0.00,0.00}{Red}, \textcolor[rgb]{0.00,0.07,1.00}{Blue}, and \textcolor[rgb]{0.00,1.00,0.00}{Green} Respectively.}
\begin{center}
\setlength{\tabcolsep}{1.3mm}{
\begin{tabular}{c|c|c|c||c|c|c}
\hline
\multirow{2}{*}{} & \multicolumn{3}{c||}{ORSSD Dataset} & \multicolumn{3}{c}{EORSSD Dataset} \\[0.5ex]
\cline{2-7}
  & $F_{\beta} \uparrow$ & MAE $\downarrow$ & $S_m \uparrow$ & $F_{\beta} \uparrow$ & MAE $\downarrow$ & $S_m \uparrow$ \\

\hline\hline
DSG \cite{DSG} & $0.6630$ & $0.1041$ & $0.7195$ & $0.5837$ & $0.1246$ & $0.6428$ \\
\hline
RRWR \cite{RRWR} & $0.5950$ & $0.1324$ & $0.6835$ & $0.4495$ & $0.1677$ & $0.5997$ \\
\hline
HDCT \cite{HDCT} & $0.5775$ & $0.1309$ & $0.6197$ & $0.5992$ & $0.1087$ & $0.5976$ \\
\hline
SMD \cite{SMD} & $0.7075$ & $0.0715$ & $0.7640$ & $0.6468$ & $0.0770$ & $0.7112$ \\
\hline
RCRR \cite{RCRR} & $0.5944$ & $0.1277$ & $0.6849$ & $0.4495$ & $0.1644$ & $0.6013$ \\
\hline
DSS \cite{DSS} & $0.7838$ & $0.0363$ & $0.8262$ & $0.7158$ & $0.0186$ & $0.7874$ \\
\hline
R3Net \cite{R3Net} & $0.7998$ & $0.0399$ & $0.8141$ & $0.7709$ & $0.0171$ & $0.8193$ \\
\hline
RADF \cite{RADF} & $0.7881$ & $0.0382$ & $0.8259$ & $0.7810$ & $0.0168$ & $0.8189$  \\
\hline
PFAN \cite{PFAN} & $0.8344$ & $0.0543$ & $0.8613$ & $0.7740$ & $0.0159$ & $0.8361$  \\
\hline
PoolNet \cite{PoolNet} & $0.7911$ & $0.0358$ & $0.8403$ & $0.7812$ & $0.0209$ & $0.8218$ \\
\hline
EGNet \cite{EGNet} & $\textcolor[rgb]{0.00,1.00,0.00}{0.8438}$ & $0.0216$ & $0.8721$ & $\textcolor[rgb]{0.00,1.00,0.00}{0.8060}$ & $\textcolor[rgb]{0.00,1.00,0.00}{0.0109}$ & $0.8602$ \\
\hline
CMC \cite{rsi5} & $0.4214$ & $0.1267$ & $0.6033$ & $0.3663$ & $0.1057$ & $0.5800$ \\
\hline
VOS \cite{rsi4} & $0.4168$ & $0.2151$ & $0.5366$ & $0.3599$ & $0.2096$ & $0.5083$ \\
\hline
SMFF \cite{SMFF} & $0.4864$ & $0.1854$ & $0.5312$ & $0.5738$ & $0.1434$ & $0.5405$ \\
\hline
LVNet \cite{LVNet} & $0.8414$ & $\textcolor[rgb]{0.00,1.00,0.00}{0.0207}$ & $\textcolor[rgb]{0.00,1.00,0.00}{0.8815}$ & $0.8051$ & $0.0145$ & $\textcolor[rgb]{0.00,1.00,0.00}{0.8645}$ \\
\hline
DAFNet-V & $\textcolor[rgb]{0.00,0.07,1.00}{0.9174}$ & $\textcolor[rgb]{0.00,0.07,1.00}{0.0125}$ & $\textcolor[rgb]{1.00,0.00,0.00}{0.9191}$ & $\textcolor[rgb]{0.00,0.07,1.00}{0.8922}$ & $\textcolor[rgb]{0.00,0.07,1.00}{0.0060}$ & $\textcolor[rgb]{0.00,0.07,1.00}{0.9167}$ \\
\hline
DAFNet-R & $\textcolor[rgb]{1.00,0.00,0.00}{0.9235}$ & $\textcolor[rgb]{1.00,0.00,0.00}{0.0106}$ & $\textcolor[rgb]{0.00,0.07,1.00}{0.9188}$ & $\textcolor[rgb]{1.00,0.00,0.00}{0.9060}$ & $\textcolor[rgb]{1.00,0.00,0.00}{0.0053}$ & $\textcolor[rgb]{1.00,0.00,0.00}{0.9185}$ \\

\hline
\end{tabular}}
\end{center}
\label{tab1}
\end{table}

\subsubsection{\textbf{Quantitative Comparison}}
For quantitative evaluations, we report the P-R curves, F-measure, MAE score, and S-measure in Fig. \ref{fig6} and Table \ref{tab1}. In addition to the VGG16 backbone, we also provide the results of the proposed DAFNet under the state-of-the-art Res2Net-50 backbone \cite{Res2Net}. From the P-R curves shown in Fig. \ref{fig6}, our DAFNet under the VGG16 backbone (\ie, the magenta solid line) is higher than all comparison methods. With the more powerful backbone (\ie, Res2Net-50), our method (\ie, the red solid line) achieves the highest precision of the whole P-R curves on these two testing datasets with remarkable margin compared with other methods.

In order to more intuitively compare the performance of different methods, the quantitative measures of different methods are reported in Table \ref{tab1}. We can draw the conclusions from the Table \ref{tab1} that are consistent with the P-R curves. For the unsupervised methods including the methods for optical RSIs, SMD \cite{SMD} method achieves the best performance in all the measurements on both two datasets. For the deep learning based method for NSIs with optical RSIs retraining, the performances of these methods are significantly better than the unsupervised methods, in which the EGNet method \cite{EGNet} achieves the best performance, and its F-measure reaches $0.8438$ on the ORSSD dataset and $0.8060$ on the EORSSD dataset. The LVNet \cite{LVNet} and EGNet \cite{EGNet} methods achieve comparable performance, but the LVNet method has obvious advantages in terms of S-measure. This also indirectly illustrates the necessity of designing a SOD solution specifically for optical RSIs. Thanks to the model design of this paper, our proposed method obtains the best performance under all quantitative measures, and has a large performance gain compared with other methods. Taking the DAFNet-V (\ie, the DAFNet under the VGG16 backbone) result as an example, compared with the \emph{\textbf{second best method}}, on the ORSSD dataset, the percentage gain reaches $8.7\%$ in terms of the F-measure, $39.6\%$ in terms of the MAE score, and $4.3\%$ in terms of the S-measure. On the EORSSD dataset, our method achieves a more significant percentage gain compared with the second best method, \ie, the percentage gain of the F-measure reaches $10.7\%$, MAE score achieves $45.0\%$, and S-measure obtains $6.0\%$. Moreover, with the powerful Res2Net-50 backbone, our method (\ie, DAFNet-R) achieves more competitive performance. All these measures demonstrate the superiority and effectiveness of the proposed method.
\begin{table}[!t]
\scriptsize
\renewcommand\arraystretch{1.2}
	\caption{F-measure Evaluation based on Category Attribute on the EORSSD Dataset.}
	\begin{center}
		\setlength{\tabcolsep}{0.2mm}{
			\begin{tabular}{ c | c c c c c c c c c }
				\hline
				 & Aircraft & Building & Car & Island & Road & Ship & Water & Non & Other \\
				\hline\hline
				LVNet \cite{LVNet} & $0.8118$ & $0.7776$ & $0.7328$ & $0.9017$ & $0.7478$ & $0.8191$ & $0.8443$ & $0.9088$ & $0.8374$ \\
				DAFNet-V           & $0.8606$ & $0.9648$ & $0.8916$ & $0.9457$ & $0.8609$ & $0.8810$ & $0.9292$ & $0.9474$ & $0.9468$ \\
				\hline
		\end{tabular}}
	\end{center}
	\label{tab2}
\end{table}

\begin{table}[!t]
\renewcommand\arraystretch{1.2}
	\caption{F-measure Evaluation based on Quantity Attribute on the EORSSD Dataset.}
	\begin{center}
		\setlength{\tabcolsep}{1mm}{
			\begin{tabular}{ c | c c c c c c }
				\hline
				 & 0 & 1 & 2 & 3 & 4 & $\ge{5}$ \\
				\hline\hline
				LVNet \cite{LVNet} & $0.9088$ & $0.8147$ & $0.7727$ & $0.7932$ & $0.8235$ & $0.8192$ \\
				DAFNet-V           & $0.9474$ & $0.9170$ & $0.8746$ & $0.8886$ & $0.9055$ & $0.8848$ \\
				\hline
		\end{tabular}}
	\end{center}
	\label{tab3}
\end{table}

\begin{table}[!t]
\renewcommand\arraystretch{1.2}
	\caption{F-measure Evaluation based on Size Attribute on the EORSSD Dataset.}
	\begin{center}
		\setlength{\tabcolsep}{0.6mm}{
			\begin{tabular}{ c | c c c c c c }
				\hline
				 & $\le{1}$\% & $1 \sim 10\%$ & $10 \sim 20\%$ & $20 \sim 30\%$ & $30 \sim 40\%$ & $\ge{40\%}$ \\
				\hline\hline
				LVNet \cite{LVNet} & $0.7727$ & $0.8166$ & $0.8681$ & $0.8752$ & $0.9019$ & $0.8598$ \\
				DAFNet-V           & $0.8692$ & $0.9233$ & $0.9403$ & $0.9428$ & $0.9519$ & $0.9633$ \\
				\hline
		\end{tabular}}
	\end{center}
	\label{tab4}
\end{table}

\subsubsection{\textbf{Attribute-based Study}}
We conduct the attribute-based study to help understand the characteristics and advantages of our model under different conditions, including category-based, quantity-based, and size-based studies.

We can draw some useful conclusions from Tables \ref{tab2}-\ref{tab4}. Apparently, our method achieves uniformly larger F-measure than its competitor LVNet \cite{LVNet} under various fine-grained testing conditions. From category-based study of our method, we can observe that the detection performance of ``aircraft'' and ``road'' objects is relatively lower compared with other categories, but is still much better than the LVNet \cite{LVNet}. The aircraft targets captured in our dataset usually show diverse appearance and suffer from complex surrounding contexts, and the road targets have long-narrow shapes and typically cover a very large scope, which significantly increases the detection difficulty. From the quantity-based study, it can be observed that the increase in the number of salient objects also leads to the increased difficulty of detection, while our method can still greatly outperform the LVNet \cite{LVNet}. From the size-based study, our method consistently shows obvious superiority over the LVNet \cite{LVNet}, especially when dealing with very tiny ($\le 1\%$) or very large objects ($\ge 40\%$). In general, despite the proposed DAFNet has significantly boosted the detection performance, there is still a lot of room for further improvement in accurately detecting the tiny objects.

\begin{figure}[!t]
\centering
\centerline{\includegraphics[width=1\linewidth]{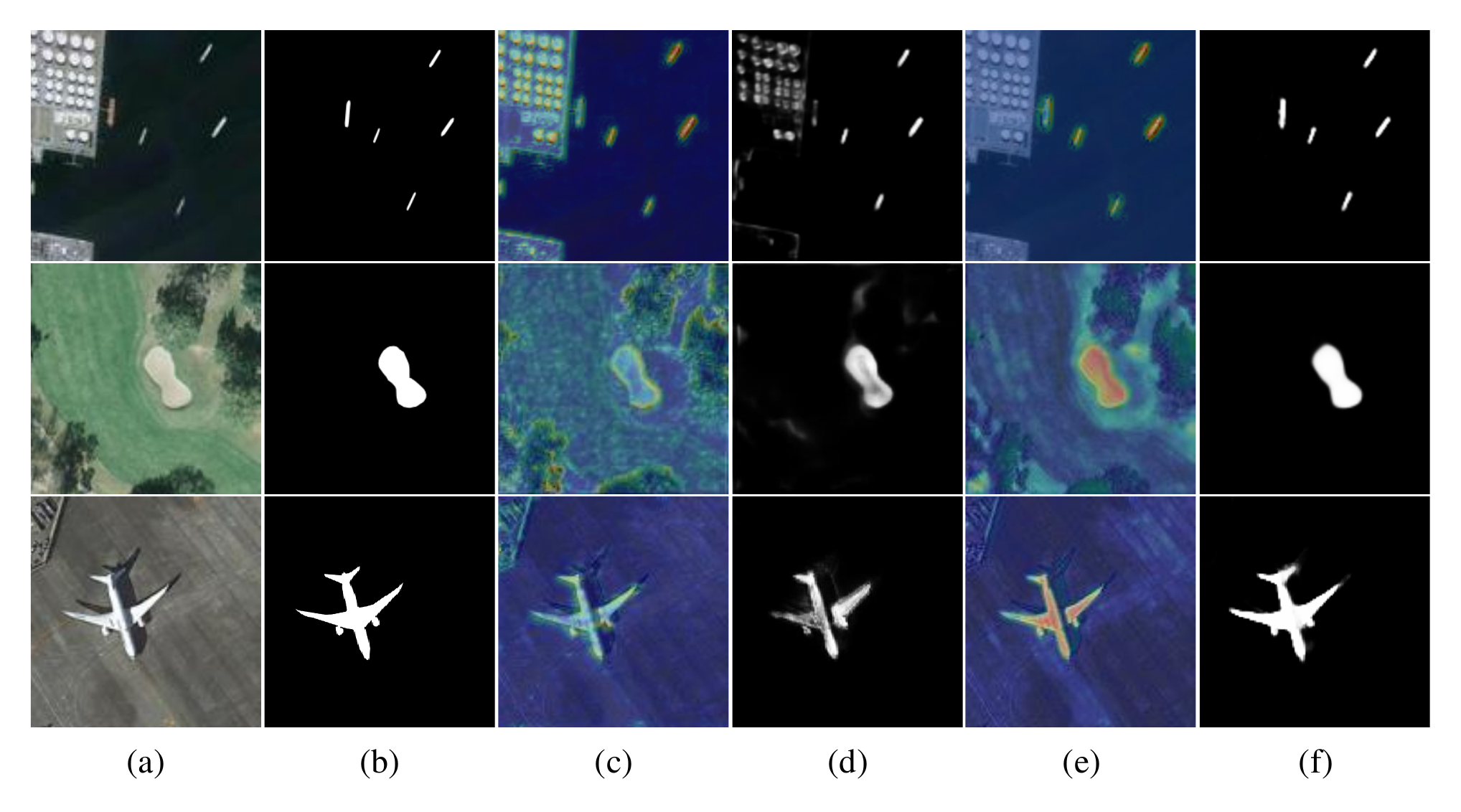}}
\caption{Visualization results of ablation study. (a) Optical RSIs. (b) Ground truths. (c)-(d) Feature and saliency maps produced by baseline network. (e)-(f) Feature and saliency maps produced by DAFNet.}
\label{fig7}
\end{figure}
\subsection{Ablation Study}

\begin{figure}[!t]
	\centering
	\centerline{\includegraphics[width=8.7cm,height=4cm]{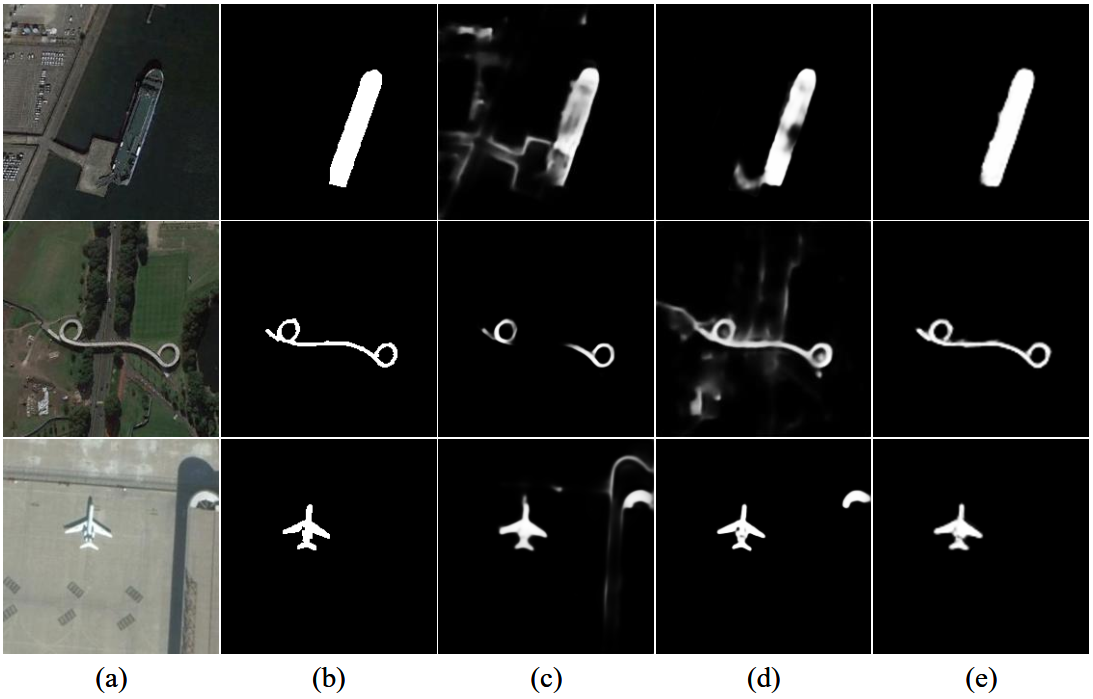}}
	\caption{Results obtained by progressively adding the GCA and DAF components to the baseline model. (a)-(b) Input RSIs and ground truths. (c) Results of Baseline. (d) Results of Baseline+GCA. (e) Results of Baseline+GCA+DAF (the full DAFNet).
	\label{gca}}
\end{figure}

There are three key modules within the DAFNet, \ie, Global Feature Aggregation (GFA), Cascaded Pyramid Attention (CPA), and Dense Attention Fluid (DAF). Accordingly, we conduct ablation experiments to evaluate the necessity and contribution of each module. The baseline model shares identical decoder architecture and optimization strategies with the full DAFNet, but it solely preserves the original backbone feature encoder. To intuitively illustrate the effectiveness of our proposed attention-driven feature enhancing strategy, we provide some feature visualizations in Fig. \ref{fig7}. As visible, compared with the baseline network, our proposed DAFNet can suppress background redundancies and produce more discriminative features, and further generate more complete saliency map with clear background. We also provide some visual evolution results by gradually adding the GCA and DAF components to the baseline model. As shown in Fig. \ref{gca}, after applying the GCA module, the model can capture more similar patterns in global image context (\eg, the long-distance bridge in the second image), and it can also suppress some background interference (\eg, the platform area on the left in the first image). Then, by further introducing the DAF module, low-level and high-level attentive guidance information are sufficiently fused to further suppress the background (\eg, the building in the third image) and extract more integral object structures (\eg, the ship in the first image).

The quantitative results of different variants are reported in Table \ref{tab5}. Note that although our baseline model shows the worst performance in all three metrics, it is still a rather powerful model compared with many of the comparison methods involved in Table \ref{tab1}. In order to encode the global contextual information in the feature learning, the GFA module is introduced and brings performance improvement, \ie, the F-measure is improved from $0.8391$ to $0.8504$ with the percentage gain of $1.3\%$, and the S-measure is boosted from $0.8432$ to $0.8661$ with the percentage gain of $2.7\%$. Then, in order to more comprehensively model multi-scale pyramid features to address the scale variation of objects in optical RSIs, we design the CPA module in a cascaded manner. The CPA module contributes to relative gains (\wrt, GFA) of $2.8\%$ in terms of F-measure, $1.1\%$ in terms of S-measure, as well as decline of $15.3\%$ in terms of MAE score. Finally, the DAF structure is used to associate multiple levels of attention modules to obtain a more comprehensive flow of the attentive information, which in turn forms our DAFNet architecture. Compared with other variants, the complete DAFNet architecture achieves the best performance, \ie, the percentage gain reaches $6.3\%$ in terms of F-measure against the baseline model, $52.0\%$ in terms of MAE, and $8.7\%$ in terms of S-measure, which demonstrates that our proposed framework benefits from all key components.

\begin{table}[!t]
\renewcommand\arraystretch{1.2}
\caption{Quantitative Evaluation of Ablation Studies on the Testing Subset of EORSSD Dataset.}
\begin{center}
\setlength{\tabcolsep}{2mm}{
\begin{tabular}{ c c c c | c c c c }
\hline
Baseline & GFA & CPA & DAF & $F_{\beta}$ & MAE & $S_m$ \\
\hline\hline
\checkmark & & & & $0.8391$ & $0.0125$ & $0.8432$ \\
\checkmark & \checkmark & & & $0.8504$ & $0.0098$ & $0.8661$ \\
\checkmark & \checkmark & \checkmark & & $0.8742$ & $0.0083$ & $0.8760$ \\
\checkmark & \checkmark & \checkmark & \checkmark & $0.8922$ & $0.0060$ & $0.9167$ \\
\hline
\end{tabular}}
\end{center}
\label{tab5}
\end{table}

\begin{figure}[!t]
	\centering
	\centerline{\includegraphics[width=1\linewidth]{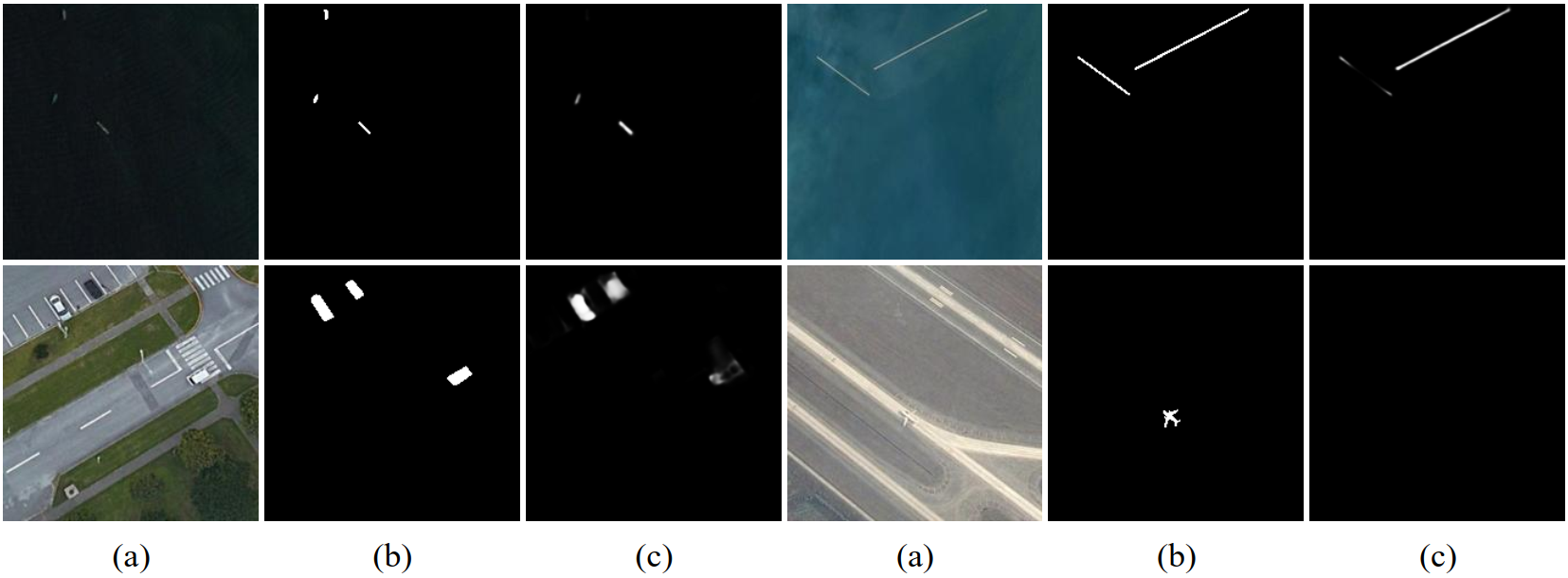}}
	\caption{Some failure examples. (a) Optical RSIs. (b) Ground truths. (c) Saliency maps produced by DAFNet.}
	\label{fig8}
\end{figure}
\subsection{Failure Cases}
We discuss some failure cases of our DAFNet in Fig. \ref{fig8}. For some very challenging examples, our method still cannot obtain perfect results.

(1) It is still challenging for our method to deal with tiny objects. For example, in the first row of Fig. \ref{fig8}, the ship in the upper left corner of the left image cannot be detected, and the thin line on the left side of the right image cannot be completely detected. The main reason is that we only use a relatively small input size due to limited GPU budget. Thus, the features of tiny salient objects would be completely lost during the down-scaling process, which hinders the successful detection. A direct solution to this problem could be exploiting more computation resources to enable the training and testing of larger-size inputs. However, it would be more promising to explore more memory-efficient global-context aware attention operations for building long-range semantic dependencies.

(2) It is still challenging to well distinguish the salient objects from the surrounding contexts with high appearance similarity. For example, in the second row of Fig. \ref{fig8}, the white car and the airplane are largely overwhelmed by the contextual texture and color information, and hence are missed by the detector. To address this issue, more attempts should be made to develop better learning strategies for the understanding of abstract semantic information in the task of saliency detection.

\section{Conclusion} \label{sec6}
This paper focuses on salient object detection in optical remote sensing images and proposes an end-to-end encoder-decoder framework dubbed as DAFNet, in which attention mechanism is incorporated to guide the feature learning. Benefiting from the attention fluid structure, our DAFNet learns to integrate low-level attention cues into the generation of high-level attention maps in deeper layers. Moreover, we investigate the global context-aware attention mechanism to encode long-range pixel dependencies and explicitly exploit global contextual information. In addition, we construct a new large-scale optical RSI benchmark dataset for SOD with pixel-wise saliency annotations. Extensive experiments and ablation studies demonstrate the effectiveness of the proposed DAFNet architecture. In the future, designing more robust SOD models with stronger generalization capabilities to achieve a win-win situation for both NSIs and optical RSIs is a promising direction for further researches.

\section{Acknowledgments} \label{sec7}
Thanks to all anonymous reviewers who helped improve this paper. This work was completed by Qijian Zhang as an intern at the Institute of Information Science, Beijing Jiaotong University.

\par
\ifCLASSOPTIONcaptionsoff
  \newpage
\fi
{
\bibliographystyle{IEEEtran}
\bibliography{ref}

\begin{thebibliography}{10}
\providecommand{\url}[1]{#1}
\csname url@samestyle\endcsname
\providecommand{\newblock}{\relax}
\providecommand{\bibinfo}[2]{#2}
\providecommand{\BIBentrySTDinterwordspacing}{\spaceskip=0pt\relax}
\providecommand{\BIBentryALTinterwordstretchfactor}{4}
\providecommand{\BIBentryALTinterwordspacing}{\spaceskip=\fontdimen2\font plus
\BIBentryALTinterwordstretchfactor\fontdimen3\font minus
  \fontdimen4\font\relax}
\providecommand{\BIBforeignlanguage}[2]{{%
\expandafter\ifx\csname l@#1\endcsname\relax
\typeout{** WARNING: IEEEtran.bst: No hyphenation pattern has been}%
\typeout{** loaded for the language `#1'. Using the pattern for}%
\typeout{** the default language instead.}%
\else
\language=\csname l@#1\endcsname
\fi
#2}}
\providecommand{\BIBdecl}{\relax}
\BIBdecl

\bibitem{wwg_review}
W.~Wang, Q.~Lai, H.~Fu, J.~Shen, and H.~Ling, ``Salient object detection in the
  deep learning era: An in-depth survey,'' \emph{arXiv preprint
  arXiv:1904.09146}, 2019.

\bibitem{REVIEW}
R.~Cong, J.~Lei, H.~Fu, M.-M. Cheng, W.~Lin, and Q.~Huang, ``Review of visual
  saliency detectioin with comprehensive information,'' \emph{IEEE Trans.
  Circuits Syst. Video Technol}, vol.~29, no.~10, pp. 2941--2959, 2019.

\bibitem{cmm_review}
A.~Borji, M.-M. Cheng, Q.~Hou, H.~Jiang, and J.~Li, ``Salient object detection:
  A survey,'' \emph{Computational Visual Media}, vol.~5, no.~2, pp. 117--150,
  2019.

\bibitem{Response_R4}
A.~Borji and L.~Itti, ``State-of-the-art in visual attention modeling,''
  \emph{IEEE Trans. Pattern Anal. Mach. Intell.}, vol.~35, no.~1, pp. 185--207,
  2012.

\bibitem{SGS}
D.~Zhao, J.~Wang, J.~Shi, and Z.~Jiang, ``Sparsity-guided saliency detection
  for remote sensing images,'' \emph{J. Applied Remote Sens.}, vol.~9, pp.
  1--14, 2015.

\bibitem{LVNet}
C.~Li, R.~Cong, J.~Hou, S.~Zhang, Y.~Qian, and S.~Kwong, ``Nested network with
  two-stream pyramid for salient object detection in optical remote sensing
  images,'' \emph{IEEE Trans. Geosci. Remote Sens.}, vol.~57, no.~11, pp.
  9156--9166, 2019.

\bibitem{Response_R9}
L.~Zhang and J.~Zhang, ``A new saliency-driven fusion method based on complex
  wavelet transform for remote sensing images,'' \emph{IEEE Geosci. Remote
  Sens. Lett.}, vol.~14, no.~12, pp. 2433--2437, 2017.

\bibitem{Response_R11}
X.~Gong, Z.~Xie, Y.~Liu, X.~Shi, and Z.~Zheng, ``Deep salient feature based
  anti-noise transfer network for scene classification of remote sensing
  imagery,'' \emph{Remote Sens.}, vol.~10, no.~3, p. 410, 2018.

\bibitem{Response_R17}
W.~Diao, X.~Sun, X.~Zheng, F.~Dou, H.~Wang, and K.~Fu, ``Efficient
  saliency-based object detection in remote sensing images using deep belief
  networks,'' \emph{IEEE Geosci. Remote Sens. Lett.}, vol.~13, no.~2, pp.
  137--141, 2016.

\bibitem{anamoly_detection}
X.~Kang, X.~Zhang, S.~Li, K.~Li, J.~Li, and J.~A. Benediktsson, ``Hyperspectral
  anomaly detection with attribute and edge-preserving filters,'' \emph{IEEE
  Trans. Geosci. Remote Sens.}, vol.~55, no.~10, pp. 5600--5611, 2017.

\bibitem{PFAN}
T.~Zhao and X.~Wu, ``Pyramid feature attention network for saliency
  detection,'' in \emph{Proc. CVPR}, 2019, pp. 3085--3094.

\bibitem{AttAdd1}
X.~Zhang, T.~Wang, J.~Qi, H.~Lu, and G.~Wang, ``Progressive attention guided
  recurrent network for salient object detection,'' in \emph{Proc. CVPR}, 2018,
  pp. 714--722.

\bibitem{AttAdd2}
Y.~Ji, H.~Zhang, and Q.~M.~J. Wu, ``Salient object detection via multi-scale
  attention {CNN},'' \emph{Neurocomputing}, vol. 322, pp. 130--140, 2018.

\bibitem{AttAdd4}
W.~Wang, S.~Zhao, J.~Shen, S.~C. Hoi, and A.~Borji, ``Salient object detection
  with pyramid attention and salient edges,'' in \emph{Proc. CVPR}, 2019, pp.
  1448--1457.

\bibitem{erf}
W.~Luo, Y.~Li, R.~Urtasun, and R.~Zemel, ``Understanding the effective
  receptive field in deep convolutional neural networks,'' in \emph{Proc.
  NeurIPS}, 2016, pp. 4898--4906.

\bibitem{EGNet}
J.-X. Zhao, J.-J. Liu, D.-P. Fan, Y.~Cao, J.-F. Yang, and M.-M. Cheng,
  ``{EGNet}: Edge guidance network for salient object detection,'' in
  \emph{Proc. ICCV}, 2019, pp. 8779--8788.

\bibitem{crmtc20}
R.~Cong, J.~Lei, H.~Fu, J.~Hou, Q.~Huang, and S.~Kwong, ``Going from {RGB} to
  {RGBD} saliency: A depth-guided transformation model,'' \emph{IEEE Trans.
  Cybern.}, vol.~50, no.~8, pp. 3627--3639, 2020.

\bibitem{crmtip18}
R.~Cong, J.~Lei, H.~Fu, Q.~Huang, X.~Cao, and C.~Hou, ``Co-saliency detection
  for {RGBD} images based on multi-constraint feature matching and cross label
  propagation,'' \emph{IEEE Trans. Image Process.}, vol.~27, no.~2, pp.
  568--579, 2018.

\bibitem{crmtc19}
R.~Cong, J.~Lei, H.~Fu, W.~Lin, Q.~Huang, X.~Cao, and C.~Hou, ``An iterative
  co-saliency framework for {RGBD} images,'' \emph{IEEE Trans. Cybern.},
  vol.~49, no.~1, pp. 233--246, 2019.

\bibitem{RC}
M.-M. Cheng, N.~J. Mitra, X.~Huang, P.~H.~S. Torr, and S.-M. Hu, ``Global
  contrast based salient region detection,'' \emph{IEEE Trans. Pattern Anal.
  Mach. Intell.}, vol.~37, no.~3, pp. 569--582, 2015.

\bibitem{RBD}
W.~Zhu, S.~Liang, Y.~Wei, and J.~Sun, ``Saliency optimization from robust
  background detection,'' in \emph{Proc. CVPR}, 2014, pp. 2814--2821.

\bibitem{HDCT}
J.~Kim, D.~Han, Y.~W. Tai, and J.~Kim, ``Salient region detection via
  high-dimensional color transform and local spatial support,'' \emph{IEEE
  Trans. Image Process.}, vol.~25, no.~1, pp. 9--23, 2015.

\bibitem{SMD}
H.~Peng, B.~Li, H.~Ling, W.~Hua, W.~Xiong, and S.~Maybank, ``Salient object
  detection via structured matrix decomposition,'' \emph{IEEE Trans. Pattern
  Anal. Mach. Intell.}, vol.~39, no.~4, pp. 818--832, 2017.

\bibitem{RCRR}
Y.~Yuan, C.~Li, J.~Kim, W.~Cai, and D.~D. Feng, ``Reversion correction and
  regularized random walk ranking for saliency detection,'' \emph{IEEE Trans.
  Image Process.}, vol.~27, no.~3, pp. 1311--1322, 2018.

\bibitem{RRWR}
C.~Li, Y.~Yuan, W.~Cai, Y.~Xia, and D.~Feng, ``Robust saliency detection via
  regularized random walks ranking,'' in \emph{Proc. CVPR}, 2015, pp.
  2710--2717.

\bibitem{DSG}
L.~Zhou, Z.~Yang, Z.~Zhou, and D.~Hu, ``Salient region detection using
  diffusion process on a two-layer sparse graph,'' \emph{IEEE Trans. Image
  Process.}, vol.~26, no.~12, pp. 5882--5894, 2017.

\bibitem{R3Net}
Z.~Deng, X.~Hu, L.~Zhu, X.~Xu, J.~Qin, G.~Han, and P.-A. Heng, ``R$^{3}${N}et:
  Recurrent residual refinement network for saliency detection,'' in
  \emph{Proc. IJCAI}, 2018, pp. 684--690.

\bibitem{DSS}
Q.~Hou, M.-M. Cheng, X.~Hu, A.~Borji, Z.~Tu, and P.~H. Torr, ``Deeply
  supervised salient object detection with short connections,'' \emph{IEEE
  Trans. Pattern Anal. Mach. Intell.}, vol.~41, no.~4, pp. 815--828, 2019.

\bibitem{RADF}
X.~Hu, L.~Zhu, J.~Qin, C.~W. Fu, and P.~A. Heng, ``Recurrently aggregating deep
  features for salient object detection,'' in \emph{Proc. AAAI}, 2018, pp.
  6943--6950.

\bibitem{BASNet}
X.~Qin, Z.~Zhang, C.~Huang, C.~Gao, M.~Dehghan, and M.~Jagersand, ``{BASNet}:
  Boundary-aware salient object detection,'' in \emph{Proc. CVPR}, 2019, pp.
  7479--7489.

\bibitem{R1_2_added_1}
H.~Zhou, X.~Xie, J.-H. Lai, Z.~Chen, and L.~Yang, ``Interactive two-stream
  decoder for accurate and fast saliency detection,'' in \emph{Proc. CVPR},
  2020, pp. 9141--9150.

\bibitem{R1_2_added_2}
J.~Wei, S.~Wang, Z.~Wu, C.~Su, Q.~Huang, and Q.~Tian, ``Label decoupling
  framework for salient object detection,'' in \emph{Proc. CVPR}, 2020, pp.
  13\,025--13\,034.

\bibitem{R1_2_added_4}
A.~Siris, J.~Jiao, G.~K. Tam, X.~Xie, and R.~W. Lau, ``Inferring attention
  shift ranks of objects for image saliency,'' in \emph{Proc. CVPR}, 2020, pp.
  12\,133--12\,143.

\bibitem{Response_R19}
Q.~Hou, M.-M. Cheng, X.~Hu, A.~Borji, Z.~Tu, and P.~H. Torr, ``Deeply
  supervised salient object detection with short connections,'' in \emph{Proc.
  CVPR}, 2017, pp. 3203--3212.

\bibitem{Response_R20}
T.-N. Le and A.~Sugimoto, ``Video salient object detection using spatiotemporal
  deep features,'' \emph{IEEE Trans. Image Process.}, vol.~27, no.~10, pp.
  5002--5015, 2018.

\bibitem{Response_R22}
M.~R. Abkenar, H.~Sadreazami, and M.~O. Ahmad, ``Graph-based salient object
  detection using background and foreground connectivity cues,'' in \emph{Proc.
  ISCAS}, 2019, pp. 1--5.

\bibitem{lcy2020tc}
C.~Li, R.~Cong, S.~Kwong, J.~Hou, H.~Fu, G.~Zhu, D.~Zhang, and Q.~Huang,
  ``{ASIF-Net}: Attention steered interweave fusion network for {RGB-D} salient
  object detection,'' \emph{IEEE Trans. Cybern.}, vol.~PP, no.~99, pp. 1--13,
  2020.

\bibitem{eccv20}
C.~Li, R.~Cong, Y.~Piao, Q.~Xu, and C.~C. Loy, ``{RGB-D} salient object
  detection with cross-modality modulation and selection,'' in \emph{Proc.
  ECCV}, 2020, pp. 1--17.

\bibitem{PoolNet}
J.-J. Liu, Q.~Hou, M.-M. Cheng, J.~Feng, and J.~Jiang, ``A simple pooling-based
  design for real-time salient object detection,'' in \emph{Proc. CVPR}, 2019,
  pp. 3917--3926.

\bibitem{GCPANet}
Z.~Chen, Q.~Xu, R.~Cong, and Q.~Huang, ``Global context-aware progressive
  aggregation network for salient object detection,'' in \emph{AAAI}, 2020, pp.
  10\,599--10\,606.

\bibitem{SMFF}
L.~Zhang, Y.~Liu, and J.~Zhang, ``Saliency detection based on self-adaptive
  multiple feature fusion for remote sensing images,'' \emph{Int. J. Remote
  Sens.}, vol.~40, no.~22, pp. 8270--8297, 2019.

\bibitem{lcync20}
C.~Li, R.~Cong, C.~Guo, H.~Li, C.~Zhang, F.~Zheng, and Y.~Zhao, ``A parallel
  down-up fusion network for salient object detection in optical remote sensing
  images,,'' \emph{Neurocomputing}, vol. 415, pp. 411--420, 2020.

\bibitem{rsi1}
L.~Ma, B.~Du, H.~Chen, and N.~Q. Soomro, ``Region-of-interest detection via
  superpixel-to-pixel saliency analysis for remote sensing image,'' \emph{IEEE
  Geosci. Remote Sens. Lett.}, vol.~13, no.~12, pp. 1752--1756, 2016.

\bibitem{rsi3}
E.~Li, S.~Xu, W.~Meng, and X.~Zhang, ``Building extraction from remotely sensed
  images by integrating saliency cue,'' \emph{IEEE J. Sel. Topics Appl. Earth
  Observ.}, vol.~10, no.~3, pp. 906--919, 2017.

\bibitem{rsi4}
Q.~Zhang, L.~Zhang, W.~Shi, and Y.~Liu, ``Airport extraction via complementary
  saliency analysis and saliency-oriented active contour model,'' \emph{IEEE
  Geosci. Remote Sens. Lett.}, vol.~15, no.~7, pp. 1085--1089, 2018.

\bibitem{rsi5}
Z.~Liu, D.~Zhao, Z.~Shi, and Z.~Jiang, ``Unsupervised saliency model with color
  markov chain for oil tank detection,'' \emph{Remote Sens.}, vol.~11, no.~9,
  pp. 1--18, 2019.

\bibitem{rsi6}
C.~Dong, J.~Liu, and F.~Xu, ``Ship detection in optical remote sensing images
  based on saliency and a rotation-invariant descriptor,'' \emph{Remote Sens.},
  vol.~10, no.~3, pp. 1--19, 2018.

\bibitem{vgg}
K.~Simonyan and A.~Zisserman, ``Very deep convolutional networks for
  large-scale image recognition,'' in \emph{Proc. ICLR}, 2015.

\bibitem{SE}
J.~Hu, L.~Shen, and G.~Sun, ``Squeeze-and-excitation networks,'' in \emph{Proc.
  CVPR}, 2018, pp. 7132--7141.

\bibitem{cbam}
S.~Woo, J.~Park, J.~Y. Lee, and I.~S. Kweon, ``{CBAM}: Convolutional block
  attention module,'' in \emph{Proc. ECCV}, 2018, pp. 1--19.

\bibitem{Response_R5}
T.~Liu, Z.~Yuan, J.~Sun, J.~Wang, N.~Zheng, X.~Tang, and H.-Y. Shum, ``Learning
  to detect a salient object,'' \emph{IEEE Trans. Pattern Anal. Mach. Intell.},
  vol.~33, no.~2, pp. 353--367, 2010.

\bibitem{crmtip19}
R.~Cong, J.~Lei, H.~Fu, F.~Porikli, Q.~Huang, and C.~Hou, ``Video saliency
  detection via sparsity-based reconstruction and propagation,'' \emph{IEEE
  Trans. Image Process.}, vol.~28, no.~10, pp. 4819--4831, 2019.

\bibitem{crmspl}
R.~Cong, J.~Lei, C.~Zhang, Q.~Huang, X.~Cao, and C.~Hou, ``Saliency detection
  for stereoscopic images based on depth confidence analysis and multiple cues
  fusion,'' \emph{IEEE Signal Process. Lett.}, vol.~23, no.~6, pp. 819--823,
  2016.

\bibitem{Fmeasure2}
R.~Achanta, S.~Hemami, F.~Estrada, and S.~Ssstrunk, ``Frequency-tuned salient
  region detection,'' in \emph{Proc. CVPR}, 2009, pp. 1597--1604.

\bibitem{nips20}
Q.~Zhang, R.~Cong, J.~Hou, C.~Li, and Y.~Zhao, ``{CoADNet}: Collaborative
  aggregation-and-distribution networks for co-salient object detection,'' in
  \emph{Proc. NeurIPS}, 2020.

\bibitem{crmicme}
Y.~Zhang, L.~Li, R.~Cong, X.~Guo, H.~Xu, and J.~Zhang, ``Co-saliency detection
  via hierarchical consistency measure,'' in \emph{ICME}, 2018, pp. 1--6.

\bibitem{MAE}
D.~Zhang, H.~Fu, J.~Han, A.~Borji, and X.~Li, ``A review of co-saliency
  detection algorithms: Fundamentals, applications, and challenges,'' \emph{ACM
  Trans. Intell. Syst. and Technol.}, vol.~9, no.~4, pp. 1--31, 2018.

\bibitem{crmtmm19}
R.~Cong, J.~Lei, H.~Fu, Q.~Huang, X.~Cao, and N.~Ling, ``{HSCS}: Hierarchical
  sparsity based co-saliency detection for {RGBD} images,'' \emph{IEEE Trans.
  Multimedia}, vol.~21, no.~7, pp. 1660--1671, 2019.

\bibitem{S-measure}
D.-P. Fan, M.-M. Cheng, Y.~Liu, T.~Li, and A.~Borji, ``Structure-measure: A new
  way to evaluate foreground maps,'' in \emph{Proc. ICCV}, 2017, pp.
  4548--4557.

\bibitem{adam}
D.~Kingma and J.~Ba, ``Adam: A method for stochastic optimization,''
  \emph{arXiv preprint arXiv:1412.6980}, 2017.

\bibitem{Xavier}
X.~Glorot and Y.~Bengio, ``Understanding the difficulty of training deep
  feedforward neural networks,'' in \emph{Proc. AISTATS}, 2010, pp. 249--256.

\bibitem{ohem}
A.~Shrivastava, A.~Gupta, and R.~Girshick, ``Training region-based object
  detectors with online hard example mining,'' in \emph{Proc. CVPR}, 2016, pp.
  761--769.

\bibitem{Res2Net}
S.-H. Gao, M.-M. Cheng, K.~Zhao, X.-Y. Zhang, M.-H. Yang, and P.~Torr,
  ``{Res2Net}: A new multi-scale backbone architecture,'' \emph{IEEE Trans.
  Pattern Anal. Mach. Intell.}, 2020.

\end{thebibliography}
}

\end{document}